\documentclass[10pt,journal,compsoc]{IEEEtran}
\usepackage[nocompress]{cite}

\usepackage[margin=0pt,font=small,labelfont=bf,labelsep=endash,tableposition=top]{caption}
\usepackage{epsfig}
\usepackage{graphicx}
\usepackage{amsmath}       
\usepackage{amssymb}       
\usepackage{booktabs}      
\usepackage{multirow}      
\usepackage{bm}            
\usepackage{algorithm}     
\usepackage{algpseudocode} 
\usepackage{algorithmicx}  
\usepackage{color}         
\usepackage{enumitem}      
\usepackage{arydshln}      
\usepackage{pifont}        
\usepackage{tablefootnote} 
\usepackage{subfig}        
\usepackage{microtype}     
\usepackage{url}           
\usepackage{mathtools,xspace} 
\usepackage{xcolor}        
\newcommand{\revise}[1]{#1}
\usepackage{hyperref}
\hypersetup{
    hidelinks,  
    breaklinks=true 
}
\usepackage{multicol}      
\usepackage{makecell}      

\usepackage[numbers]{natbib}

\newlength\savedwidth

\begin{document}

\title{Agentic Designer: Progressive Multi-Agent Collaboration for Structure-Aware Interior Layout Generation}

\author{Zhijing~Yang,
        Haocheng~Lin,
        Zhihua~Xu,
        Haojie~Li,
        Keze~Wang,
        Liang~Lin,~\IEEEmembership{Fellow,~IEEE},
        and~Tianshui~Chen
\thanks{Z. Yang, H. Lin and Z. Xu contributed equally to this work and share first authorship.
Z. Yang, T. Chen, and H. Lin, are with the Guangdong University of Technology, Guangzhou, China
(Emails: linhaocheng1@mails.gdut.edu.cn, tianshuichen@gmail.com, yzhj@gdut.edu.cn).
Z. Xu, K. Wang, and L. Lin are with Sun Yat-sen University, Guangzhou, China
(Email: xuzhh76@mail2.sysu.edu.cn, kezewang@gmail.com, linliang@ieee.org).
H. Li is with the South China University of Technology, Guangzhou, China
(Email: 12hjli4@gmail.com).
(Corresponding author: T. Chen).

This work was supported in part by the National Natural Science Foundation of China (NSFC) under Grant 62276283, in part by Fundamental Research Funds for the Central Universities, Sun Yat-sen University under Grant 23hytd006 and 23hytd006-2, and  in part by Guangdong Provincial High-Level Young Talent Program under Grant RL2024-151-2-11.
}
}


\markboth{IEEE Transactions on Pattern Analysis and Machine Intelligence}%
{Lin \MakeLowercase{\textit{et al.}}: Agentic Designer: Progressive Multi-Agent Collaboration for Structure-Aware Interior Layout Generation}

\IEEEtitleabstractindextext{%
\begin{abstract}
Generating realistic interior furniture layouts that strictly adhere to architectural constraints (e.g., walls, doors, and windows) remains a fundamental challenge in automated spatial design. Existing approaches, primarily based on one-shot generation using diffusion models or Large Language Models (LLMs), lack explicit mechanisms for intermediate geometric constraint verification, often resulting in structural collisions and functionally infeasible arrangements under complex room constraints. To address these challenges, we propose Agentic Designer, a progressive, multi-agent framework that formulates structure-aware interior layout generation as an iterative and constraint-verified decision process. By decomposing layout synthesis into modular stages of proposal, verification, and adjustment, the framework coordinates three specialized agents, a Generator, an Evaluator, and a Refiner, through a Progressive Consensus Mechanism. This mechanism enforces stepwise geometric validation and correction before each placement is committed, thereby preventing error accumulation. To facilitate this structure-aware paradigm and standardize evaluation, we establish InStruct, a comprehensive benchmark that integrates a dataset comprising over 18,000 high-quality, parametrically annotated samples with a novel suite of structure-centric metrics. Extensive quantitative evaluations, qualitative analyses, and user studies show that Agentic Designer significantly outperforms state-of-the-art methods, demonstrating substantial improvements in strict structural adherence and functional design coherence.
\end{abstract}

}

\maketitle

\IEEEdisplaynontitleabstractindextext

\IEEEpeerreviewmaketitle

\IEEEraisesectionheading{\section{Introduction}\label{sec:introduction}}
Structure-aware interior layout generation is introduced in this work as a new task that reformulates traditional layout design by explicitly modeling architectural elements, such as walls, doors, and windows, as structural constraints, and inferring feasible furniture placement, orientation, and scale conditioned on these constraints. With the growing interest in automated interior layout generation~\cite{wang2019planit,paschalidou2021atiss,feng2023layoutgpt,tang2024diffuscene,yang2024physcene,liang2025s}, realism and practicality have become central research concerns. Interior layout design in real-world living environments is inherently governed by architectural structures, which define feasible placement, movement affordances, and functional usability. However, most existing automated interior layout generation methods simplify rooms into homogeneous bounded regions~\cite{paschalidou2021atiss}, treating architectural elements as implicit or secondary constraints~\cite{tang2024diffuscene,feng2023layoutgpt}. This abstraction overlooks the fundamentally structure-dependent nature of furniture arrangement: doors must remain accessible, windows unobstructed, and furniture aligned with walls and openings. As a result, layouts generated under this assumption frequently violate structural feasibility and practical usability, particularly in complex room configurations.

Recent research on automated interior layout generation can be broadly categorized into two directions. The first focuses on end-to-end parametric generation, which directly synthesizes furniture layouts from given conditions using diffusion-based models or large language model (LLM) prompting~\cite{tang2024diffuscene,feng2023layoutgpt}. The second adopts a two-stage pipeline, where an intermediate representation (e.g., semantic or spatial maps) is first produced and then translated into concrete furniture parameters~\cite{wang2019planit,sun2025semlayoutdiff}. It is also worth noting that LLM-based approaches employ prompt engineering to guide general-purpose language models, rather than models specifically trained for spatial or design-oriented reasoning, to infer furniture layouts from textual cues~\cite{hu2024scenecraft,feng2023layoutgpt}. Although these approaches have significantly advanced automation in layout design, they still struggle to maintain structural alignment with architectural elements and coherent spatial relationships among furniture pieces, often resulting in unrealistic or impractical arrangements under complex room constraints.

The limitations of current methods mainly arise from their one-shot generation paradigm and insufficient data quality. Most approaches map input conditions directly to final layouts without intermediate reasoning or feedback, making it difficult to enforce structural constraints or control object-level properties such as position, orientation, and scale~\cite{chen2026learning}. The lack of efficient evaluation further limits interpretability and prevents correction of suboptimal configurations. In contrast, human designers typically adopt an iterative strategy, placing, evaluating, and refining furniture piece by piece, which naturally ensures both functional consistency and spatial coherence~\cite{tanasra2023automation,wang2025chat2layout}. In addition, existing datasets are often small in scale, lack complete structural annotations, and contain inaccurate furniture placements~\cite{chang2017matterport3d, song2017semantic, zheng2020structured3d, fu20213d}, restricting precise model training and structural alignment. Therefore, it is essential to develop a higher-quality dataset with comprehensive structural annotations and precise spatial information. Such a dataset is critical for advancing the accuracy, generalizability, and structural alignment of models in interior layout generation.

Motivated by these observations, we introduce Agentic Designer, a progressive multi-agent framework that iteratively constructs and refines interior layouts through coordinated collaboration. The framework comprises three specialized agents, a Generator, an Evaluator, and a Refiner, each responsible for proposing, assessing, and improving furniture arrangements under predefined spatial and functional constraints. To enhance consistency and interpretability in layout reasoning, it employs a structured, program-like textual representation that explicitly encodes spatial relations and design logic. Building upon this foundation, a Progressive Consensus Mechanism (PCM) orchestrates successive stages of generation, evaluation, and refinement, ensuring steady improvement in both geometric accuracy and aesthetic coherence. These design choices enable Agentic Designer to produce layouts that are structurally aligned, spatially coherent, and amenable to iterative, controllable, and interpretable refinement.

To foster the advancement of structure-aware interior layout generation and standardize its evaluation, we establish InStruct, a comprehensive benchmark specifically tailored for this domain. Beyond merely addressing data scarcity, InStruct provides a unified testbed comprising 18,853 samples (10,795 living rooms and 8058 bedrooms). Each sample is manually verified and richly annotated with architectural elements, including walls, doors, and windows, alongside precise furniture attributes. To further broaden the evaluation landscape, the benchmark additionally incorporates an adapted version of the widely used 3D-FRONT dataset~\cite{fu20213d}, reprocessed to align with our task formulation and integrated into the testing protocols. Moreover, these data resources are complemented by a novel suite of structure-centric metrics designed to quantify geometric alignment and functional compatibility with architectural elements. By integrating diverse data sources with multi-dimensional evaluation protocols, InStruct serves as a foundational platform for assessing the realism and structural feasibility of structure-aware layout generation.

Our contribution can be summarized as the following points:
\begin{itemize}
    \item 
    We propose Agentic Designer, a progressive multi-agent framework that mimics human design workflows by integrating a Generator, Evaluator, and Refiner. This architecture transforms layout generation from an opaque one-shot process into a controllable, interpretable, and self-correcting pipeline.
    \item 
    We introduce a Progressive Consensus Mechanism (PCM) to orchestrate agent collaboration. This mechanism ensures consistent refinement across design stages, allowing for precise alignment with architectural constraints while steadily optimizing spatial coherence and functionality.
    \item 
    We establish InStruct, a comprehensive benchmark tailored for structure-aware layout generation. By providing high-quality manually verified data and specialized structure-centric metrics, it addresses the critical issue of data scarcity and standardizes the evaluation of structural feasibility.
    \item 
    Experimental results show that our method consistently improves geometric accuracy and design coherence, demonstrating that agent-based collaboration effectively resolves conflicts between strict structural constraints and design flexibility. Codes, trained models, and datasets will be available at \url{https://haochengcop.github.io/Agentic-Designer/}. 
\end{itemize}

\section{Related Work}
\label{sec:related_work}

In this section, we review the literature relevant to our proposed framework, covering one-shot generative models for interior layout generation, large language models for spatial reasoning, agent-based collaborative systems, and existing interior furniture layout datasets.

\subsection{One-Shot Generative Models for Interior Layout Generation}
Early approaches to automated interior layout generation primarily relied on Convolutional Neural Networks (CNNs)~\cite{ritchie2019fast} and Generative Adversarial Networks (GANs)~\cite{yang2021indoor, tanasra2023automation, chen2025contrastive} to synthesize 2D floor plans or object arrangements via image-based generation~\cite{wang2019planit, li2019grains}. While these methods established the feasibility of data-driven design, they were often limited by grid-based resolutions and lacked object-level manipulation capabilities.

The subsequent shift towards sequence modeling established autoregressive Transformers as a dominant paradigm~\cite{paschalidou2021atiss, wang2021sceneformer}. For instance, ATISS~\cite{paschalidou2021atiss} treats layout generation as a sequence prediction task, synthesizing a sequence of furniture objects based on learned spatial distributions. More recently, the success of diffusion models in image synthesis has inspired approaches~\cite{wei2023lego, lin2024instructscene, hu2024mixed, tang2024diffuscene, sun2025semlayoutdiff} like DiffuScene~\cite{tang2024diffuscene} and SemLayoutDiff~\cite{sun2025semlayoutdiff}, which model layouts as a denoising process from Gaussian noise. These methods have significantly improved the diversity and realism of generated scenes.

Despite these improvements, continuous-space models face inherent limitations in structural adherence. By mapping inputs directly to coordinates without explicit logical reasoning, they often struggle to satisfy hard geometric constraints—such as preventing furniture from overlapping with doors or ensuring precise alignment with walls. Additionally, their black-box nature complicates interpretability, making it difficult to understand placement decisions or intervene when suboptimal layouts are produced.

\subsection{LLM-Based Spatial Reasoning and Agentic Collaboration}
The emergence of Large Language Models (LLMs) has reframed spatial planning as a semantic reasoning task. 
Recent studies~\cite{feng2023layoutgpt, ccelen2024design, fu2024anyhome, yang2024holodeck}, such as LayoutGPT, 
leverage the rich pretrained knowledge embedded in LLMs to infer object arrangements directly from textual descriptions. 
More recent approaches~\cite{hu2024scenecraft}, including SceneCraft, adopt executable code (e.g., Python scripts) as an intermediate representation~\cite{liang2022code}, 
which more naturally captures hierarchical structures and logical constraints than raw coordinates. 
However, although these code-driven approaches offer greater expressiveness, they do not necessarily guarantee structural correctness. 
Operating predominantly in a one-shot paradigm, these methods map instructions directly to layouts without intermediate verification. 
Consequently, without domain-specific adaptation or explicit geometric validation, these models frequently produce hallucinated arrangements~\cite{brohan2023can}, 
where semantically relevant furniture is placed in physically implausible configurations, such as object collisions or violations of room boundaries.

To address the limitations of one-shot generation, recent research in complex reasoning domains has increasingly explored  multi-agent systems~\cite{hong2023metagpt, shi2024mtr++, liu2024interaction, lin2025mao} and iterative refinement strategies~\cite{madaan2023self, hu2023mo, besta2025demystifying}. 
In these frameworks, distinct agents assume specialized roles, such as generators, critics, and editors, to collaboratively solve problems that are difficult for a single model to handle. While demonstrated to be effective across various reasoning-intensive domains, the application of such collaborative paradigms to interior layout generation remains underexplored. 
In practice, human designers typically follow an iterative process: proposing an initial layout, evaluating it against spatial constraints, and refining it accordingly. 
Inspired by this workflow, we introduce a Generator--Evaluator--Refiner loop that dynamically detects and corrects spatial violations during layout generation.

\subsection{Interior Layout Datasets and Benchmarks}
Existing datasets for interior layout generation generally fall into two categories: real-world reconstructions and synthetic environments. 

Real-world datasets include established benchmarks like Matterport3D~\cite{chang2017matterport3d} and ScanNet~\cite{dai2017scannet}, as well as higher-fidelity contemporary datasets like HM3D~\cite{ramakrishnan2021habitat} and ScanNet++~\cite{yeshwanth2023scannet++}. While recent advancements have significantly improved geometric resolution, these datasets fundamentally rely on unstructured surface meshes rather than parametric representations. In such environments, architectural boundaries and furniture are depicted as geometries often fused with the floor or background, suffering from occlusion and sensor artifacts. This lack of explicit, clean structural definition renders them suboptimal for layout generation tasks, as models struggle to extract precise geometric constraints from such fragmented data.

Synthetic datasets emerged to circumvent the geometric noise and incompleteness of scanned data. SUNCG~\cite{song2017semantic} was a pioneering effort but is no longer available. Subsequent works have focused on different aspects: InteriorNet~\cite{li2018interiornet} emphasizes photo-realistic rendering and physics simulation, while Structured3D~\cite{zheng2020structured3d} provides richer structural annotations such as room wireframes. Currently, 3D-FRONT~\cite{fu20213d} stands as the mainstream benchmark due to its large scale and high-quality furniture models. 
Nevertheless, despite its object diversity, 3D-FRONT is primarily optimized for visual rendering rather than structural reasoning. Its architectural elements are represented as unstructured mesh geometries rather than parametric entities. This forces current layout generation models to infer constraints from raw meshes, inevitably introducing geometric ambiguity and limiting the ability to learn precise, rule-based design logic.

To address this, we establish InStruct, a benchmark that integrates a new, structurally annotated dataset with an adapted version of 3D-FRONT. Paired with specialized metrics, it enables rigorous evaluation of structural constraints and geometric validity.

\section{Method}

\begin{figure*}[ht]
    \centering
    \includegraphics[width=0.98\linewidth]{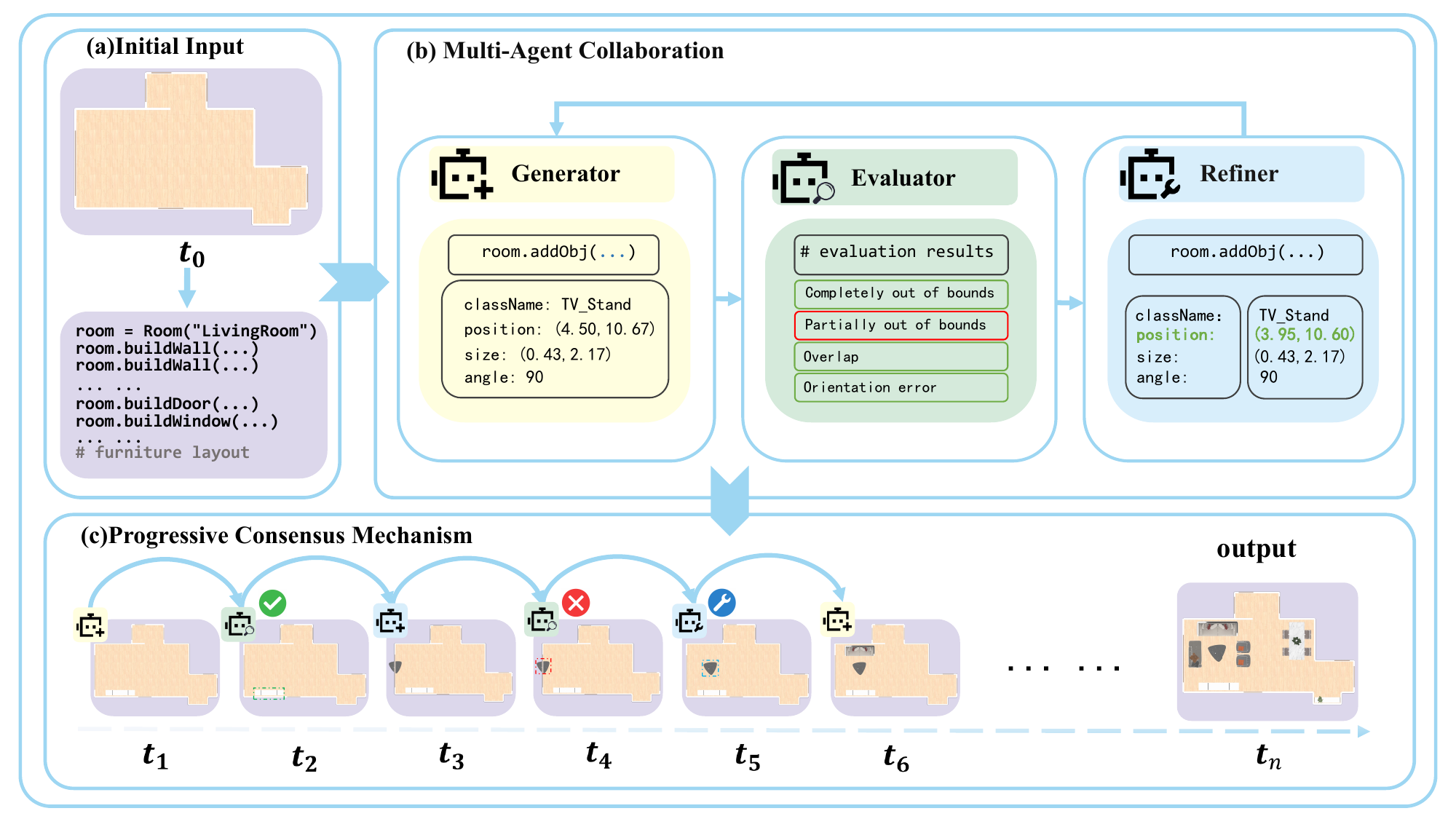}
    \vspace{-3mm}
    \caption{Framework overview of the proposed Agentic Designer.
(a) Initial Input: The architectural structure is converted into a program-like representation that encodes walls, doors, and windows as the structural context.
(b) Multi-Agent Collaboration: The layout is iteratively constructed by three collaborative agents. The Generator proposes candidate furniture objects, the Evaluator diagnoses geometric violations (e.g., boundary intrusion, collisions, orientation errors), and the Refiner performs targeted corrections to restore validity while preserving design intent.
(c) Progressive Consensus Mechanism: Through step-wise proposal, verification, and refinement, the agents progressively construct an optimal layout while minimizing error accumulation.The process repeats until the Generator outputs the termination instance ($c=\texttt{End}$).}
    \vspace{-3mm}
    \label{fig:framework}
\end{figure*}

As illustrated in Figure~\ref{fig:framework}, we propose \textbf{Agentic Designer}, a multi-agent framework that iteratively constructs interior layouts through three collaborative modules: a Generator, an Evaluator, and a Refiner. In this section, we detail the framework's components and their coordination. We begin by formulating the structure-aware layout generation task in Sec.~\ref{sec:problem}, followed by the specific definitions of each agent in Sec.~\ref{sec:agents}. Subsequently, we describe the collaborative workflow driven by the \textbf{Progressive Consensus Mechanism} in Sec.~\ref{sec:protocol}, and conclude with implementation details in Sec.~\ref{sec:implementation}.

\subsection{Problem Formulation}
\label{sec:problem}
We formulate the structure-aware interior layout generation task as inferring a layout configuration that conforms to architectural constraints defined by $\mathcal{S} = \{\mathcal{S}_{w}, \mathcal{S}_{d}, \mathcal{S}_{g}\}$, where $\mathcal{S}_{w}$, $\mathcal{S}_{d}$, and $\mathcal{S}_{g}$ represent the sets of walls, doors, and windows, respectively. Each structural element $s \in \mathcal{S}$ is modeled as an axis-aligned quadrilateral $s = ( v^{(1)}, v^{(2)}, v^{(3)}, v^{(4)} )$ with $v^{(k)} \in \mathbb{R}^2$, providing a unified geometric basis for constraint enforcement. The goal is to generate an ordered sequence of furniture instances $\mathcal{O} = (o_1, o_2, \dots, o_N)$ that maintains spatial coherence within these boundaries. Each object $o_i$ is defined by a tuple $o_i = (c_i, p_i, d_i, \theta_i)$, where $c_i \in \mathcal{C}$ denotes the semantic category, $p_i \in \mathbb{R}^2$ is the center position, $d_i \in \mathbb{R}^4$ represents the 2D bounding box, and $\theta_i \in [0, 2\pi)$ indicates the orientation.

\subsection{Multi-Agent Collaboration}
\label{sec:agents}
To enable progressive, controllable, and structure-aware layout construction, we propose Agentic Designer, a multi-agent framework that models interior design as an iterative decision-making process rather than a one-shot prediction. Mirroring the workflow of human designers, the framework decomposes layout synthesis into modular stages of proposal, verification, and adjustment, orchestrated by three collaborative agents: a \textbf{Generator} that incrementally proposes furniture instances, an \textbf{Evaluator} that assesses geometric plausibility against architectural constraints, and a \textbf{Refiner} that resolves detected inconsistencies through targeted corrections. This collaborative design ensures steady improvement in geometric accuracy and allows for interpretable reasoning throughout the generation process.

\subsubsection{Generator Agent}
The Generator is responsible for the incremental proposal of furniture objects. At step $i$, conditioned on the architectural structure $\mathcal{S}$ and the history of placed objects $\mathcal{O}_{<i}$, it predicts a candidate object $\hat{o}_i = G(\mathcal{S}, \mathcal{O}_{<i})$, where $\hat{o}_i = (c_i, p_i, d_i, \theta_i)$. To autonomously conclude the layout expansion, the Generator is designed to produce a specific object instance assigned with the semantic category $c_i=\texttt{End}$. This instance serves as an explicit termination indicator, signifying that the current layout is complete and requires no further additions.

\subsubsection{Evaluator Agent}
\revise{The Evaluator functions as a learned, LLM-based geometric consistency checker, verifying whether each proposed object complies with architectural constraints. It is fine-tuned with role-specific supervision to identify geometric violations from structured layout representations. Given a candidate $\hat{o}_i$, the structural context $\mathcal{S}$, and the previously accepted objects $\mathcal{O}_{<i}$, it produces a structured diagnostic report with Boolean judgments across four predefined violation types: \emph{partial boundary intrusion}, \emph{complete boundary violation}, \emph{object collision}, and \emph{orientation misalignment}. We formally denote this four-way diagnosis as $e_i = E(\hat{o}_i, \mathcal{S}, \mathcal{O}_{<i})$, where $e_i \in \{0,1\}^{4}$ and each entry corresponds to one violation type in the above order. A value of $1$ indicates the presence of the corresponding violation, while $0$ indicates its absence. This type-specific diagnostic feedback is then provided to the Refiner for targeted correction.}

\subsubsection{Refiner Agent}
\revise{Conditioned on the diagnostic feedback from the Evaluator, the Refiner performs object-level geometric refinement of infeasible proposals. It outputs a refined object $\hat{o}_i' = R(\hat{o}_i, e_i, \mathcal{S}, \mathcal{O}_{<i})$, with updated attributes $\hat{o}_i' = (c_i, p_i', d_i', \theta_i')$, where the semantic category $c_i$ remains unchanged from the original proposal. The refinement is applied only to fine-grained geometric attributes, including position, dimensions, and orientation. These local corrections aim to resolve the specific violations indicated by $e_i$ while preserving the original design intent of the Generator.}

\subsection{Progressive Consensus Mechanism}
\label{sec:protocol}
To effectively orchestrate the specialized agents defined above, we introduce the Progressive Consensus Mechanism, a coordinated workflow designed to bridge the gap between static generation and dynamic design reasoning. Unlike conventional end-to-end approaches where early structural violations propagate and amplify throughout the generation process, our mechanism mimics the meticulous, step-by-step workflow of human designers. By decomposing the layout synthesis into a sequence of verify-and-refine steps, we ensure that each design decision is validated or corrected before serving as a basis for subsequent steps, thereby reducing the accumulation of local structural errors before they propagate to subsequent generation steps. The detailed execution flow is depicted in Figure~\ref{fig:framework}(c) and proceeds as follows:

\begin{algorithm}[t]
\caption{Progressive Consensus Mechanism}
\label{alg:pcp}
\begin{algorithmic}[1]
    \Require Architectural Structure $\mathcal{S}$
    \Ensure Generated Furniture Layout $\mathcal{O}$
    \State \textbf{Initialize:} $\mathcal{O} \gets \emptyset$, $i \gets 1$
    \While{True}
        \State \textbf{Step 2: Conditional Proposal}
        \State $\hat{o}_i \gets \text{Generator}(\mathcal{S}, \mathcal{O})$
        \If{$\hat{o}_i.c = \texttt{End}$} \textbf{break} \EndIf
        \State \textbf{Step 3: Evaluation and Refinement}
        \State $e_i \gets \text{Evaluator}(\hat{o}_i, \mathcal{S}, \mathcal{O})$
        \If{$e_i = \mathbf{0}$} 
            \State $o_i \gets \hat{o}_i$
        \Else 
            \State $o_i \gets \text{Refiner}(\hat{o}_i, e_i, \mathcal{S}, \mathcal{O})$
        \EndIf
        \State \textbf{Step 4: Update State}
        \State $\mathcal{O} \gets \mathcal{O} \cup \{o_i\}$, $i \gets i + 1$
    \EndWhile
    \State \Return $\mathcal{O}$
\end{algorithmic}
\end{algorithm}

\noindent
\textbf{Step 1: Structural Context Initialization.}
\revise{The process commences with the stage shown in Figure~\ref{fig:framework}(a)}, where the raw architectural geometry $\mathcal{S}$ is encoded into a structured, program-like textual representation. This explicit coding serves as the global context, grounding all subsequent agent interactions in a precise coordinate system defined by the room's boundary elements (walls, doors, and windows).

\noindent
\textbf{Step 2: Conditional Proposal.}
In iteration $i$, the \textbf{Generator Agent} receives the structural context and the history of placed furniture $\mathcal{O}_{<i}$. It samples a candidate object $\hat{o}_i$ from the learned conditional distribution:
\begin{equation}
    \hat{o}_i \sim P_{\theta}(o \mid \mathcal{S}, \mathcal{O}_{<i}).    
\end{equation}

This sampling strategy enables the model to explore diverse, plausible configurations rather than being limited to deterministic outputs.

\noindent
\textbf{Step 3: Evaluation and Refinement.}
The candidate $\hat{o}_i$ is immediately assessed by the \textbf{Evaluator Agent} to determine its geometric validity.
\begin{itemize}
    \item \textit{Validation:} If the Evaluator detects no violations ($e_i = \mathbf{0}$), the object is directly accepted as $o_i$ and appended to the layout.
    \item \textit{Correction:} If violations are detected ($e_i \neq \mathbf{0}$), the \textbf{Refiner Agent} is triggered. Conditioned on the Evaluator's specific diagnostic feedback, the Refiner applies targeted transformations to resolve the inconsistencies, yielding a corrected object $\hat{o}_i'$. 
\end{itemize}

\noindent
\textbf{Step 4: State Update and Iteration.}
After the object is accepted or refined, the system updates the state to $\mathcal{O}_{\le i}$. The workflow loops back to Step 2 to generate the next object. This progressive cycle concludes only when the Generator autonomously predicts the termination instance ($c = \texttt{End}$). This signal implies that the agent perceives the current spatial arrangement as functionally complete and structurally saturated, requiring no further additions.

A critical advantage of the Progressive Consensus Mechanism is its ability to enforce a "virtuous cycle" of accuracy. In standard autoregressive generation, the probability of a valid layout $\mathcal{O}$ is given by $P(\mathcal{O}) = \prod_{i} P(o_i \mid \hat{\mathcal{O}}_{<i})$. If an early object $\hat{o}_k$ in history $\hat{\mathcal{O}}_{<i}$ is flawed (e.g., overlapping a wall), it corrupts the spatial context for all subsequent steps $i > k$, leading to cascading errors.

In contrast, our mechanism encourages that at every step $i$, the conditioning history $\mathcal{O}_{<i}$ is maintained in a structurally verified state with respect to the validity constraint set $\mathbb{V}$:
\begin{equation}
    \forall o_k \in \mathcal{O}_{<i}, \quad E(o_k, \mathcal{S}, \mathcal{O}_{<k}) \approx \mathbf{0}.
\end{equation}
This indicates that the Generator operates on a cleaner and more reliable context, where previously detected structural errors have been mitigated through evaluation and refinement. Consequently, the prediction precision for the current step can be enhanced by the improved structural integrity of the previous steps:
\begin{equation}
    P_{\mathrm{valid}}(o_i \mid \mathcal{O}_{<i}^{\mathrm{verified}}) > P_{\mathrm{valid}}(o_i  \mid \mathcal{O}_{<i}^{\mathrm{noisy}}).
\end{equation}

This property helps the layout quality improve progressively, as reliable context cues (e.g., clear available space, aligned orientation) guide the placement of subsequent furniture, leading to more globally coherent arrangements with fewer collisions and structural violations.

\subsection{Implementation Details}
\label{sec:implementation}
In this section, we detail the implementation of Agentic Designer, starting with the unified model architecture and structured input representation. We then describe the role-specific data construction strategies for each agent, clarify the bounded refinement strategy used during inference, and finally present the training protocols, including data augmentation and hyperparameter settings.

\subsubsection{Model Architecture and Input Representation.}
\revise{The Generator, Evaluator, and Refiner are all built on the Qwen2.5-Coder-7B~\cite{hui2024qwen2} backbone, but are independently trained with role-specific data and task objectives. During inference, they are deployed as separate modules and communicate strictly through structured input/output prompts, including candidate furniture proposals, diagnostic violation reports, and corrected furniture proposals. The choice of a code-specialized language model is motivated by our structured, program-like layout representation, which requires precise syntactic and logical reasoning~\cite{zhang2025system, plaat2025multi, liu2025eliciting}. To facilitate distinct roles, we design specific system prompts for each agent (detailed in \textbf{Appendix C}). }All inputs, including the room structure, historical layout context, and target object attributes, are formatted as structured code sequences, enabling the model to leverage its code generation capabilities for spatial reasoning.

\subsubsection{Agent-Specific Data Construction.}
Although the three agents adopt the same pretrained backbone architecture, they are independently fine-tuned using role-specific training pairs and task objectives:
\begin{itemize}
    \item \textbf{Generator Agent (Sequential Modeling):} To enable logical layout expansion, we organize the furniture instances in each training sample based on semantic importance and frequency (e.g., placing large distinct items like beds or sofas before accessory items)~\cite{wang2025chat2layout}. The training data is formatted as an autoregressive task: given the room structure and the sequence of preceding furniture $\mathcal{O}_{<i}$, the model is trained to predict the attributes of the current furniture $o_i$.
    
    \item \textbf{Evaluator Agent (Error Discrimination):} We synthesize training data for geometric verification by injecting controlled noise into ground-truth layouts. For a valid furniture object $o_i$, we apply random perturbations (translation and rotation) to generate a "noisy" counterpart $\tilde{o}_i$ that violates specific constraints. The input consists of the valid history $\mathcal{O}_{<i}$ and the perturbed object $\tilde{o}_i$, while the target output is structured diagnostic feedback indicating the presence or absence of each predefined violation type. This trains the agent to discriminate between valid and invalid configurations.
    
    \item \textbf{Refiner Agent (Denoising and Correction):} To empower the correction phase of the Progressive Consensus Mechanism, the Refiner is trained on pairs of invalid and valid states derived from the Evaluator's data generation process. The input comprises the valid history $\mathcal{O}_{<i}$, the perturbed object $\tilde{o}_i$, and the corresponding structured diagnostic feedback. The target output is the original, ground-truth object $o_i$. This setup essentially models refinement as a denoising task, teaching the agent to restore geometric validity based on diagnostic feedback.
\end{itemize}

\subsubsection{Data Augmentation.}
To enhance model robustness and rotational invariance, we apply geometric augmentation to the dataset. Each room layout is rotated by $0^\circ, 90^\circ, 180^\circ,$ and $270^\circ$, effectively quadrupling the size of the training set. This ensures that the agents learn to handle architectural constraints regardless of the room's global orientation.

\subsubsection{Training Setup.}
We fine-tune the models using Low-Rank Adaptation (LoRA)~\cite{hu2022lora} to ensure parameter efficiency. The LoRA rank is set to $r=32$ with an alpha scaling factor of $\alpha=64$. We use the AdamW optimizer~\cite{loshchilov2017decoupled} with a learning rate of $1\text{e-}4$. All experiments are conducted on 4 NVIDIA RTX 4090 GPUs, and the training process is monitored to prevent overfitting.

\section{The InStruct Benchmark}

To provide a unified and rigorous evaluation framework for \textit{structure-aware interior layout generation}, we introduce the \textbf{InStruct Benchmark}, which focuses on explicit architectural constraints and spatial coherence.
\revise{The benchmark integrates our high-quality \textbf{InStruct} dataset with a carefully adapted version of \textbf{3D-FRONT}. 
For 3D-FRONT, we transform raw object-centric mesh representations into explicit parametric architectural elements, including walls, doors, and windows, to ensure consistency with the proposed task formulation.}
The \textbf{InStruct} dataset is split into training, validation, and test sets with a ratio of \textbf{9:1:1}.

The remainder of this section is organized as follows.
We first describe the dataset construction pipeline and the adaptation of 3D-FRONT in Section~\ref{sec:construction}.
Section~\ref{sec:data_stats} then presents the dataset statistics and structural diversity.
Section~\ref{sec:baseline} introduces the baseline methods included in the benchmark.
Finally, Section~\ref{sec:metrics} and Section~\ref{sec:protocols} define the evaluation metrics and unified evaluation protocols.

\subsection{Dataset Construction and Analysis}
\label{sec:dataset}

\subsubsection{Data Construction Pipeline}
\label{sec:construction}

To ensure the scalability and reliability of the benchmark, we construct the InStruct dataset from a large collection of real-world residential floor plans through a rigorously designed \textit{iterative human-in-the-loop} pipeline. A primary challenge arises from the substantial visual heterogeneity of these real-world floor plans, which span diverse drawing styles such as CAD renderings, presentation drawings, and hand-drawn sketches. To address this issue, we adopt a style-adaptive construction strategy that enables robust structural perception across heterogeneous inputs.

Specifically, the pipeline integrates automated annotation with progressive human verification in an iterative manner. Floor plans are first categorized by visual style. \revise{They are then processed using specialized detectors tailored to each style to produce preliminary structural annotations for architectural elements, such as walls, doors, and windows.} High-confidence predictions are subsequently verified and corrected by human annotators, and the refined annotations are iteratively fed back to improve subsequent rounds of automated processing. Following convergence, a final global audit is conducted to eliminate residual inconsistencies and enforce annotation uniformity. \revise{All verified detections are then projected into real-world coordinates using the floor plan scale. This produces standardized, structure-aware layout annotations for benchmarking and quantitative evaluation.}

For fair and unified benchmarking, we additionally adapt the widely used 3D-FRONT~\cite{fu20213d} dataset into the same structure-aware parametric representation. Room boundaries are reconstructed from raw mesh geometry and projected onto 2D floor plans. Architectural openings are identified by consolidating multiple mesh components and resolving ambiguous or inconsistent mesh-level semantics, where a single category name may correspond to either doors or windows. \revise{To enable explicit structural reasoning, the extracted floor boundaries are converted into a unified wall representation with standardized thickness. Door and window instances are then geometrically aligned with the corresponding wall segments.} The resulting layouts form clean, topology-consistent structural annotations that are directly comparable to InStruct. Further details on the construction of the InStruct dataset and the associated annotation quality-control procedures are provided in \textbf{Appendix A}. 

\subsubsection{Dataset Statistics and Diversity}
\label{sec:data_stats}
We analyze the comprehensive characteristics of InStruct from three key perspectives: scale, structural complexity, and semantic diversity.
\noindent\textbf{Overall Scale.} 
The \textbf{InStruct} dataset comprises a total of 18,853 high-quality interior layouts, specifically consisting of 10,795 living rooms and 8,058 bedrooms. These scenes are populated with over 181,880 parametrically annotated furniture instances. This substantial scale provides a rich foundation for training data-hungry generative models, ensuring sufficient diversity to capture complex spatial dependencies between furniture objects and architectural structures.

\begin{figure}[htbp] 
    \centering
    \includegraphics[width=\linewidth]{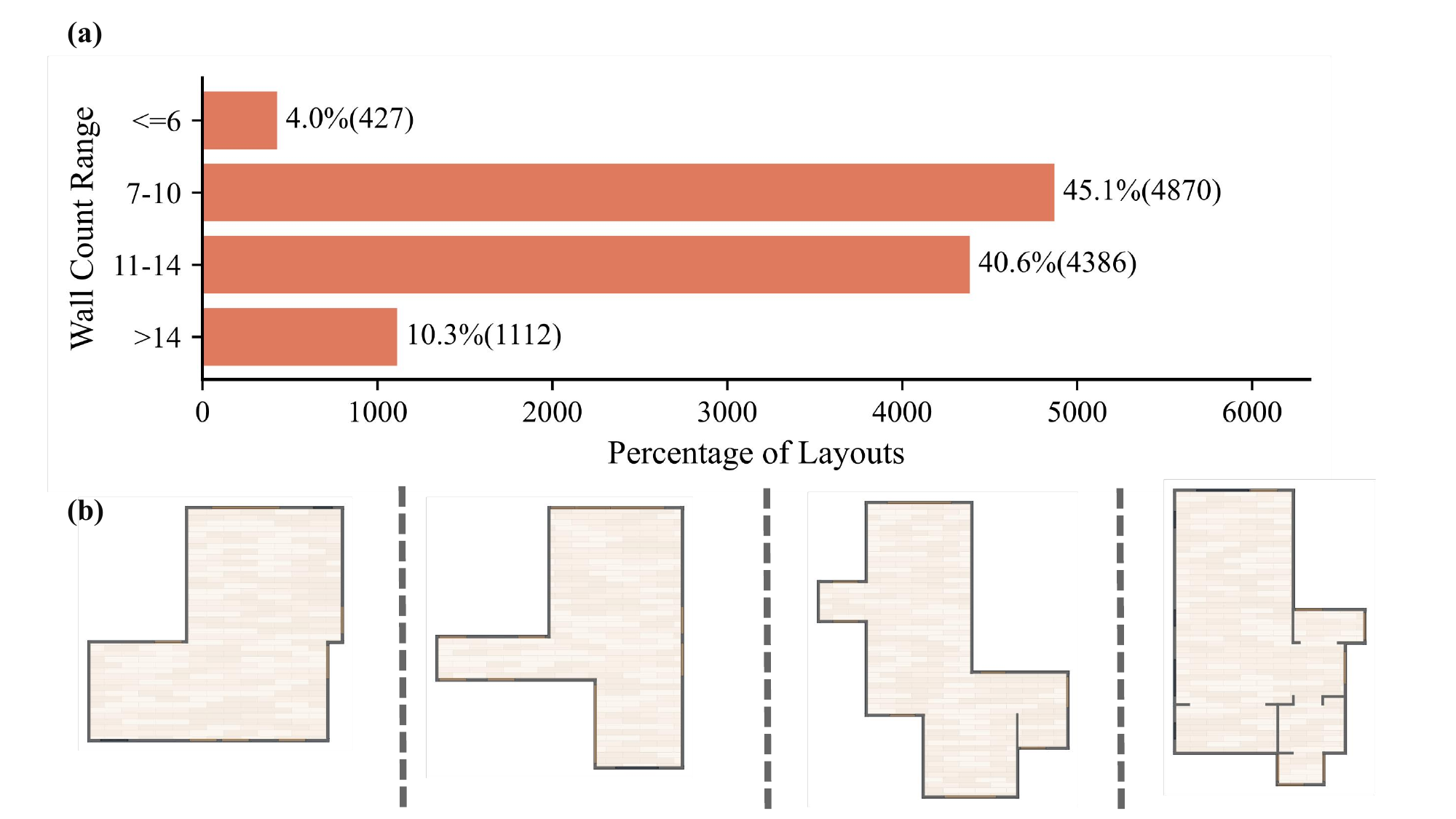}
    \caption{Wall-segment distribution for living-room layouts in the \textbf{InStruct} dataset. Most living rooms exhibit complex, non-rectangular boundaries, with the majority containing more than six wall segments.}
    \label{fig:wall_dist_liv}
\end{figure}

\begin{figure}[htbp] 
    \centering
    \includegraphics[width=\linewidth]{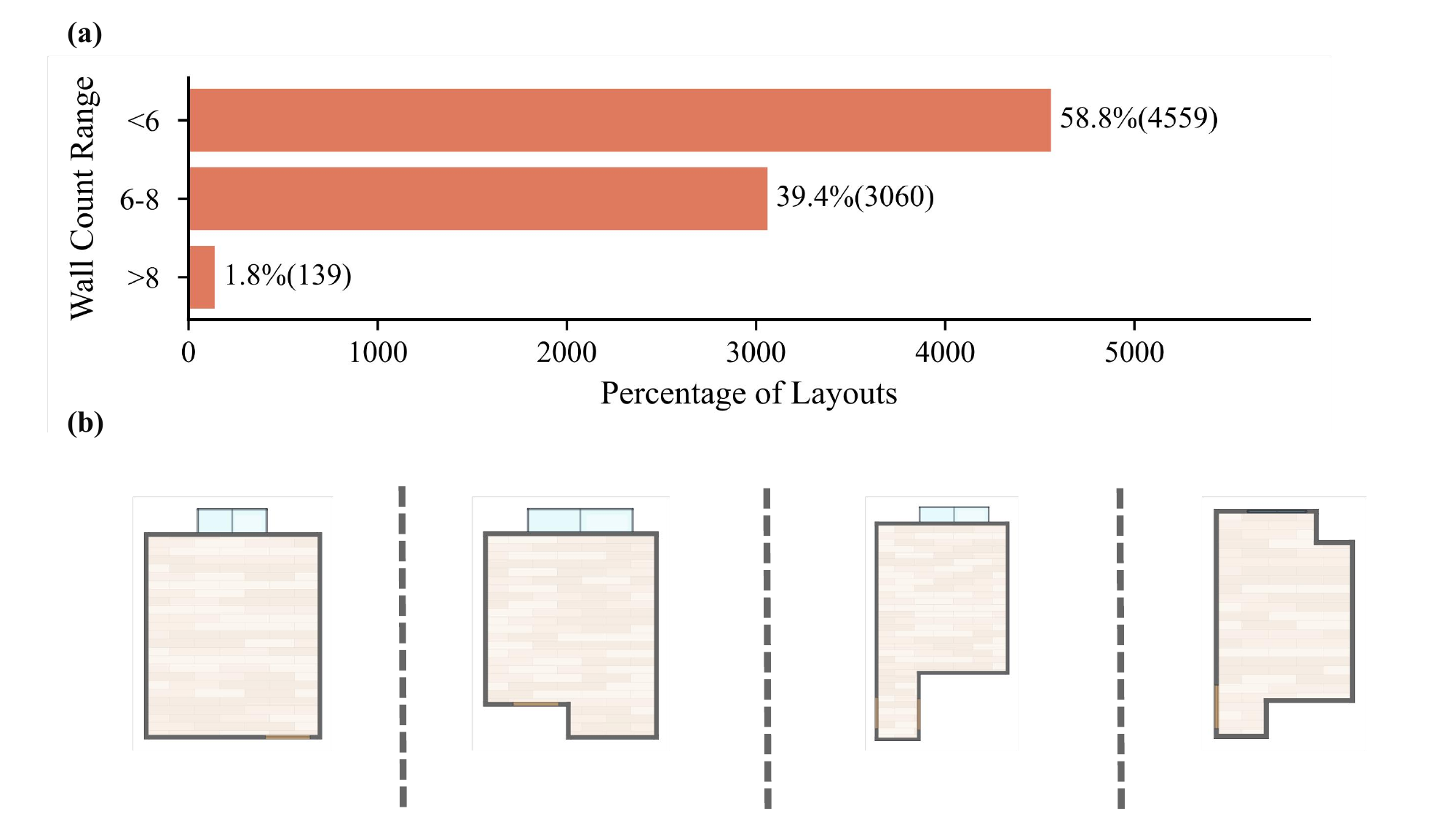}
    \caption{Wall-segment distribution for bedroom layouts. Compared with living rooms, bedrooms show a higher proportion of simple geometries, though a substantial number still contain six or more segments.}
    \label{fig:wall_dist_bed}
\end{figure}

\noindent\textbf{Structural Complexity.}
A key characteristic of InStruct is the pronounced topological diversity of its room structures. Unlike many existing synthetic datasets that are dominated by simple rectangular geometries, InStruct incorporates a substantial proportion of complex, non-rectangular room layouts. As shown in Fig.~\ref{fig:wall_dist_liv} and Fig.~\ref{fig:wall_dist_bed}, the majority of layouts consist of more than six wall segments, particularly in living rooms, while a significant portion of bedrooms exhibits similar structural complexity. In addition to topological complexity, the dataset further exhibits highly intricate boundary conditions characterized by frequent architectural openings (i.e., doors and windows). \revise{These openings densely interrupt wall boundaries across both room types. As a result, they impose rigorous geometric constraints and reframe layout generation as a strict constraint satisfaction problem rather than a relatively open-space arrangement.}

\noindent\textbf{Semantic Diversity.}
InStruct encompasses a broad semantic space with 15 distinct functional furniture categories, covering common interior elements across sleeping, dining, storage, and seating activities.
Beyond category coverage, semantic diversity in InStruct is primarily manifested at the room level through rich multi-object compositions, where multiple heterogeneous furniture items jointly form a single layout.
Such room layouts are typically populated with functionally diverse and non-sparse object sets.
On average, living rooms contain 13.42 furniture instances per scene, while bedrooms contain 4.80 instances. 



\subsubsection{Comparison with 3D-FRONT Dataset}
\label{sec:data_comparison}
\revise{We compare InStruct with 3D-FRONT~\cite{fu20213d}, a widely used large-scale indoor scene dataset. 
The comparison highlights differences in structural representation, geometric consistency, and suitability for structure-aware interior layout generation.}
We evaluate both datasets using the unified metrics defined in Sec.~\ref{sec:metrics}, which jointly characterize scene richness, space utilization, and structural consistency. As summarized in Table~\ref{tab:dataset_comparison}, InStruct consistently demonstrates substantially improved structural validity across both living-room and bedroom layouts. In particular, layouts derived from 3D-FRONT exhibit high rates of structural violations and object collisions, reflecting frequent wall penetrations and overlaps inherited from raw mesh-based placements~\cite{fu20213d}. In contrast, InStruct markedly reduces these violations, validating the effectiveness of its explicit parametric annotation and structural standardization pipeline.

\begin{table}[ht]
\centering
\caption{\textbf{Statistical comparison between InStruct and 3D-FRONT.}
Higher ANF indicates greater scene richness. Lower values are better for OR, SVR, BVC, and CS (pairwise IoU).}
\label{tab:dataset_comparison}
\setlength{\tabcolsep}{6pt}
\renewcommand{\arraystretch}{1.25}
\resizebox{\linewidth}{!}{
\begin{tabular}{llccccc}
\toprule
\textbf{Room Type} & \textbf{Dataset} & \textbf{ANF} $\uparrow$ & \textbf{OR (\%)} $\uparrow$ & \textbf{SVR (\%)} $\downarrow$ & \textbf{BVC} $\downarrow$ & \textbf{CS (IoU)} $\downarrow$ \\
\midrule
\multirow{2}{*}{Living Room} 
& 3D-FRONT & 10.59 & 31.51 & 64.85 & 1.28 & 46.8 \\
& \textbf{InStruct} & \textbf{13.41} & \textbf{22.51} & \textbf{12.37} & \textbf{0.26} & \textbf{4.7} \\
\midrule
\multirow{2}{*}{Bedroom} 
& 3D-FRONT & 4.34 & 47.98 & 73.45 & 1.34 & 90.0 \\
& \textbf{InStruct} & \textbf{4.87} & \textbf{36.90} & \textbf{28.86} & \textbf{0.35} & \textbf{1.2} \\
\bottomrule
\end{tabular}
}
\end{table}

Importantly, these gains in structural consistency are achieved without compromising scene richness. \revise{InStruct maintains higher or comparable furniture density across room types, such as 13.41 versus 10.59 objects per living room. 
This indicates that InStruct can support dense yet physically plausible multi-object arrangements under stricter geometric constraints.}

\begin{figure*}[ht]
    \centering
    \includegraphics[width=0.98\linewidth]{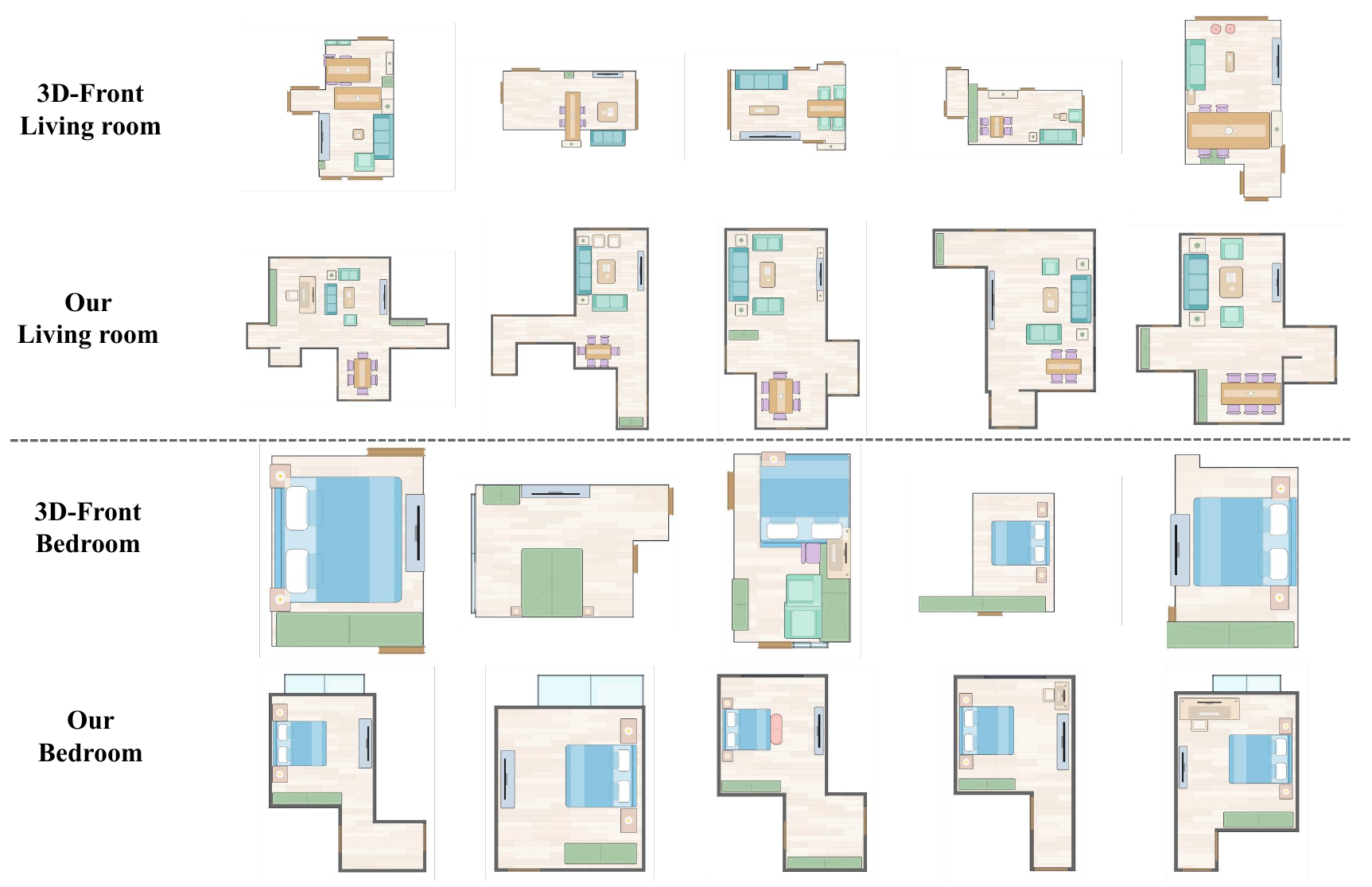}
    \vspace{-3mm}
    \caption{Visual comparison between 3D-FRONT and InStruct for living-room and bedroom layouts. 
InStruct exhibits more realistic floor-plan geometries and more coherent spatial organization, 
with furniture placements that align more consistently with architectural structures.}
    \vspace{-3mm}
    \label{fig:dataSetCompare}
\end{figure*}

Figure~\ref{fig:dataSetCompare} provides representative visual comparisons between the two datasets. InStruct layouts exhibit floor-plan geometries that more closely resemble real residential designs, including non-rectangular boundaries, natural indentations, and corridor-like extensions. Furniture placements also show stronger alignment with architectural structures and more balanced spatial organization, preserving circulation space and functional zoning. By contrast, layouts projected from 3D-FRONT tend to appear more rectangular and visually congested, with weaker alignment between furniture objects and structural boundaries.

These differences stem from a fundamental divergence in dataset representation and design objectives. InStruct explicitly models architectural elements—such as walls, doors, and windows—as parametric entities, rendering structural constraints directly accessible for reasoning and optimization. In contrast, 3D-FRONT adopts an object-centric representation in which architectural context is encoded implicitly in raw meshes with inconsistent semantics, making reliable extraction of boundary constraints challenging. Consequently, while 3D-FRONT is well suited for object-centric scene synthesis and perceptual modeling, InStruct provides a more appropriate foundation for structure-aware and constraint-driven interior layout generation.

\subsection{Baseline Methods.}
\label{sec:baseline}
We select DiffuScene~\cite{tang2024diffuscene} and SemLayout~\cite{sun2025semlayoutdiff} as representative state-of-the-art baselines for interior layout generation. DiffuScene formulates scene synthesis as a continuous diffusion process over an unordered set of object attributes, where furniture placements are generated without explicitly modeling architectural structures. In contrast, SemLayout represents room geometry, doors, and windows as a 2D semantic layout map and generates furniture arrangements by reasoning over semantic relations on a discretized grid.

\subsection{Evaluation Metrics}
\label{sec:metrics}
To thoroughly evaluate the generated layouts, the InStruct benchmark establishes a multi-dimensional evaluation suite comprising quantitative geometric metrics and subjective perceptual metrics. This combination ensures a holistic assessment of both mathematical correctness and human-centric design quality. 

\subsubsection{Quantitative Metrics}
These metrics focus on scene complexity and geometric validity, assessing the richness of the generated content and the model's adherence to structural constraints: 
\begin{itemize}
\item \textbf{Average Number of Furniture (ANF):} The arithmetic mean of furniture instances per generated scene. This metric indicates the model's capacity to synthesize spatially rich and complex layouts, rather than sparse or empty rooms.

\item \textbf{Occupancy Ratio (OR):} This measures the average ratio of the total furniture area to the room area. It reflects the spatial density and usage efficiency of the generated layouts.

\item \textbf{Scene Violation Rate (SVR):} The percentage of generated layouts containing at least one furniture instance that penetrates architectural boundaries (e. g., walls). This serves as a strict, scene-level indicator of structural adherence.

\item \textbf{Boundary Violation Count (BVC):} The average number of individual furniture instances per scene that violate boundary constraints. Unlike SVR, this metric provides a fine-grained assessment of local geometric errors, quantifying the severity of structural misalignment. 

\item \textbf{Collision Score (CS):} To evaluate object-level physical plausibility, we compute the average pairwise Intersection-over-Union (IoU) between all furniture instances. We scale the result by a factor of $10^4$ for readability.  A lower score indicates fewer physical collisions and better spatial separation.
\end{itemize}

\subsubsection{Perceptual Metrics (Subjective)}
To capture the functional usability and aesthetic quality that quantitative formulas may overlook, we define three perceptual dimensions for human evaluation based on a 5-point Likert scale (ranging from 1: Poor to 5: Excellent):
\begin{itemize}
\item \textbf{Circulation Rationality (CR):} Assesses whether the layout preserves smooth and unobstructed movement flows. This includes ensuring that furniture does not block doorways or impede access to functional zones. 

\item \textbf{Location Rationality (LR):} Evaluates the functional logic of furniture placement and its consistency with room geometry.  Key considerations include adherence to boundary constraints, prevention of invalid overlaps, and reasonable spatial relationships (e.g., a TV stand positioned opposite a sofa).

\item \textbf{Proportional Rationality (PR):} Measures the visual balance and scale harmony of the layout, examining whether furniture dimensions are appropriate relative to the room area and to surrounding objects. 
\end{itemize}

\subsection{Evaluation Protocols}
\label{sec:protocols}

\textbf{Method-Specific Input Adaptation.}
To ensure a fair comparison across methods with heterogeneous input representations, we explicitly account for their distinct architectural modeling paradigms in the evaluation protocol. DiffuScene~\cite{tang2024diffuscene} generates layouts by denoising continuous object attributes (e.g., size and position) from a binary room-boundary mask, whereas SemLayout~\cite{sun2025semlayoutdiff} operates on a 2D semantic grid that encodes room boundaries, doors, and windows. In contrast, our Agentic Designer consumes a parametric, program-like representation of architectural structures.  
\revise{For evaluation on the InStruct test set, we perform method-specific input adaptation. 
Specifically, we project the parametric annotations of walls, doors, and windows into the exact input formats required by each baseline.} All models are evaluated using their original configurations, with DiffuScene and SemLayout instantiated from the official checkpoints released by the authors. This setup allows each method to operate within its intended representation space while enabling direct comparison on the same test scenarios.

\textbf{Cross-Dataset Generalization.}
Beyond InStruct, we further examine cross-dataset generalization on 3D-FRONT~\cite{fu20213d}, the dataset originally used by DiffuScene and SemLayout. We randomly sample 300 living rooms and 300 bedrooms from the curated subset of the adapted 3D-FRONT dataset, and apply the corresponding input adaptation for DiffuScene, SemLayout, and Agentic Designer, respectively, while keeping all models fixed without additional fine-tuning. This bidirectional evaluation protocol facilitates a comprehensive assessment of structure-aware layout generation across different data formulations.

\section{Experiments}
\label{sec:experiments}
In this section, we comprehensively evaluate Agentic Designer for structure-aware interior layout generation. We first compare our framework against state-of-the-art methods quantitatively on the InStruct and 3D-FRONT datasets in Section~\ref{sec:quantitative_analysis}, followed by qualitative visual comparisons in Section~\ref{sec:qualitative_analysis} and a user study assessing perceptual realism and functional usability in Section~\ref{sec:user_study}. To validate our system design, we conduct ablation studies examining the core multi-agent collaboration and iterative refinement mechanisms in Section~\ref{sec:ablation_study}. Finally, we provide in-depth analyses of the Refiner's convergence behavior under varying iteration limits in Section~\ref{sec:convergence_analysis} and the inference latency introduced by the progressive multi-agent loop in Section~\ref{sec:latency_analysis}.

\subsection{Quantitative Analysis}
\label{sec:quantitative_analysis}

We present quantitative comparisons on the InStruct test set and the 3D-FRONT test set in Table~\ref{tab:quantitative_sota} and Table~\ref{tab:quant_3dfront}, respectively. The results show that Agentic Designer consistently outperforms state-of-the-art baselines, particularly in terms of strict geometric validity and structural adherence.

\subsubsection{Performance on InStruct}
As shown in Table~\ref{tab:quantitative_sota}, existing one-shot generation methods (DiffuScene and SemLayout) struggle under the rigorous structural constraints of the InStruct dataset, exhibiting extremely high Scene Violation Rates (SVR $> 88\%$) and Collision Scores, which indicate difficulties in precisely aligning furniture with complex wall geometries and boundary conditions. In contrast, Agentic Designer achieves a substantial reduction in structural errors, reducing the SVR for living rooms from over 90\% (baselines) to 16.85\%, while lowering the Boundary Violation Count (BVC) by an order of magnitude (from $\sim5.0$ to 0.30). Importantly, this improvement in validity does not come at the cost of scene simplicity: our method generates layouts with a higher Average Number of Furniture (ANF: 13.16 vs. 10.23) and a more realistic Occupancy Ratio (OR). These results indicate that the collaborative agent workflow can effectively resolve conflicts in dense arrangements, producing layouts that are both semantically rich and geometrically plausible, whereas baselines often result in locally congested and structurally inconsistent configurations.

\subsubsection{Generalization on 3D-FRONT}
Table~\ref{tab:quant_3dfront} further evaluates the generalization capability of our method on the 3D-FRONT dataset. Even though baselines such as DiffuScene were originally developed under this data distribution, they still exhibit substantial structural violations (e.g., an SVR of 86.00\% for living rooms). When adapted to the same inputs, Agentic Designer consistently demonstrates stronger structural reasoning, achieving the lowest violation rates (SVR: 14.81\% for living rooms and 8.33\% for bedrooms) along with minimal boundary violations. While preserving a comparable or higher furniture count (ANF), our method maintains a competitive Collision Score (CS). Overall, these results suggest that the proposed Progressive Consensus Mechanism is robust across different data distributions, effectively mitigating the hallucination of physically impossible placements through iterative agent coordination rather than one-shot generation.

\begin{table*}[htbp]
\centering
\caption{\textbf{Quantitative comparison with state-of-the-art methods on the InStruct test set.}
We evaluate geometric validity and scene complexity across Living Room and Bedroom scenarios.
\textbf{ANF}: Average Number of Furniture, \textbf{OR}: Occupancy Ratio, \textbf{SVR}: Scene Violation Rate, \textbf{BVC}: Boundary Violation Count, \textbf{CS}: Collision Score.
$\uparrow$ indicates higher is better, and $\downarrow$ indicates lower is better. Our method significantly outperforms baselines in structural adherence and collision avoidance.}
\label{tab:quantitative_sota}
\setlength{\tabcolsep}{8pt} 
\renewcommand{\arraystretch}{1.2}
\resizebox{\linewidth}{!}{ 
\begin{tabular}{llccccc}
\toprule
\textbf{Room Type} & \textbf{Method} & \textbf{ANF} $\uparrow$ & \textbf{OR (\%)} $\uparrow$ & \textbf{SVR (\%)} $\downarrow$ & \textbf{BVC} $\downarrow$ & \textbf{CS} $\downarrow$ \\
\midrule
\multirow{3}{*}{Living Room} 
& DiffuScene~\cite{tang2024diffuscene} & 10.23 & 3.32 & 92.90 & 5.35 & 18.00 \\
& SemLayout~\cite{sun2025semlayoutdiff} & 9.69 & 11.85 & 91.69 & 4.85 & 7.71 \\
& \textbf{Agentic Designer (Ours)} & \textbf{13.16} & \textbf{21.55} & \textbf{16.85} & \textbf{0.30} & \textbf{3.27} \\ 
\midrule
\multirow{3}{*}{Bedroom} 
& DiffuScene~\cite{tang2024diffuscene} & 4.15 & 3.38 & 95.14 & 3.37 & 31.70 \\
& SemLayout~\cite{sun2025semlayoutdiff} & 3.63 & 39.35 & 88.90 & 2.04 & 30.62 \\
& \textbf{Agentic Designer (Ours)} & \textbf{4.78} & \textbf{35.73} & \textbf{7.14} & \textbf{0.14} & \textbf{3.10} \\ 
\bottomrule
\end{tabular}
}
\end{table*}

\begin{table*}[htbp]
\centering
\caption{\textbf{Quantitative comparison on the 3D-FRONT test set.} 
Results are reported on 300 living rooms and 300 bedrooms randomly sampled from the 3D-FRONT test split. 
Higher ANF indicates richer layouts, while lower values are better for OR, SVR, BVC, and CS.}
\label{tab:quant_3dfront}
\setlength{\tabcolsep}{8pt} 
\renewcommand{\arraystretch}{1.2} 
\resizebox{\linewidth}{!}{ 
\begin{tabular}{llccccc}
\toprule
\textbf{Room Type} & \textbf{Method} 
& \textbf{ANF} $\uparrow$ 
& \textbf{OR (\%)} $\uparrow$ 
& \textbf{SVR (\%)} $\downarrow$ 
& \textbf{BVC} $\downarrow$ 
& \textbf{CS (IoU)} $\downarrow$ \\ 
\midrule
\multirow{3}{*}{Living Room} 
& DiffuScene~\cite{tang2024diffuscene} 
& 10.19 & 5.13 & 86.00 & 4.40 & 20.77 \\
& SemLayout~\cite{sun2025semlayoutdiff} 
& 9.82 & 19.28 & 93.67 & 4.01 & 8.67 \\
& \textbf{Agentic Designer (Ours)}
& \textbf{10.86} & \textbf{23.67} & \textbf{14.81} & \textbf{0.27} & \textbf{9.72} \\ 
\midrule
\multirow{3}{*}{Bedroom} 
& DiffuScene~\cite{tang2024diffuscene} 
& 4.28 & 11.76 & 76.33 & 1.77 & 36.96 \\
& SemLayout~\cite{sun2025semlayoutdiff} 
& 3.37 & 40.96 & 88.33 & 1.96 & 24.46 \\
& \textbf{Agentic Designer (Ours)}
& \textbf{4.49} & \textbf{36.06} & \textbf{8.33} & \textbf{0.08} & \textbf{9.88} \\ 
\bottomrule
\end{tabular}}
\end{table*}

\subsection{Qualitative Analysis}
\label{sec:qualitative_analysis}

\begin{figure*}[htbp]
    \centering
    \includegraphics[width=\textwidth]{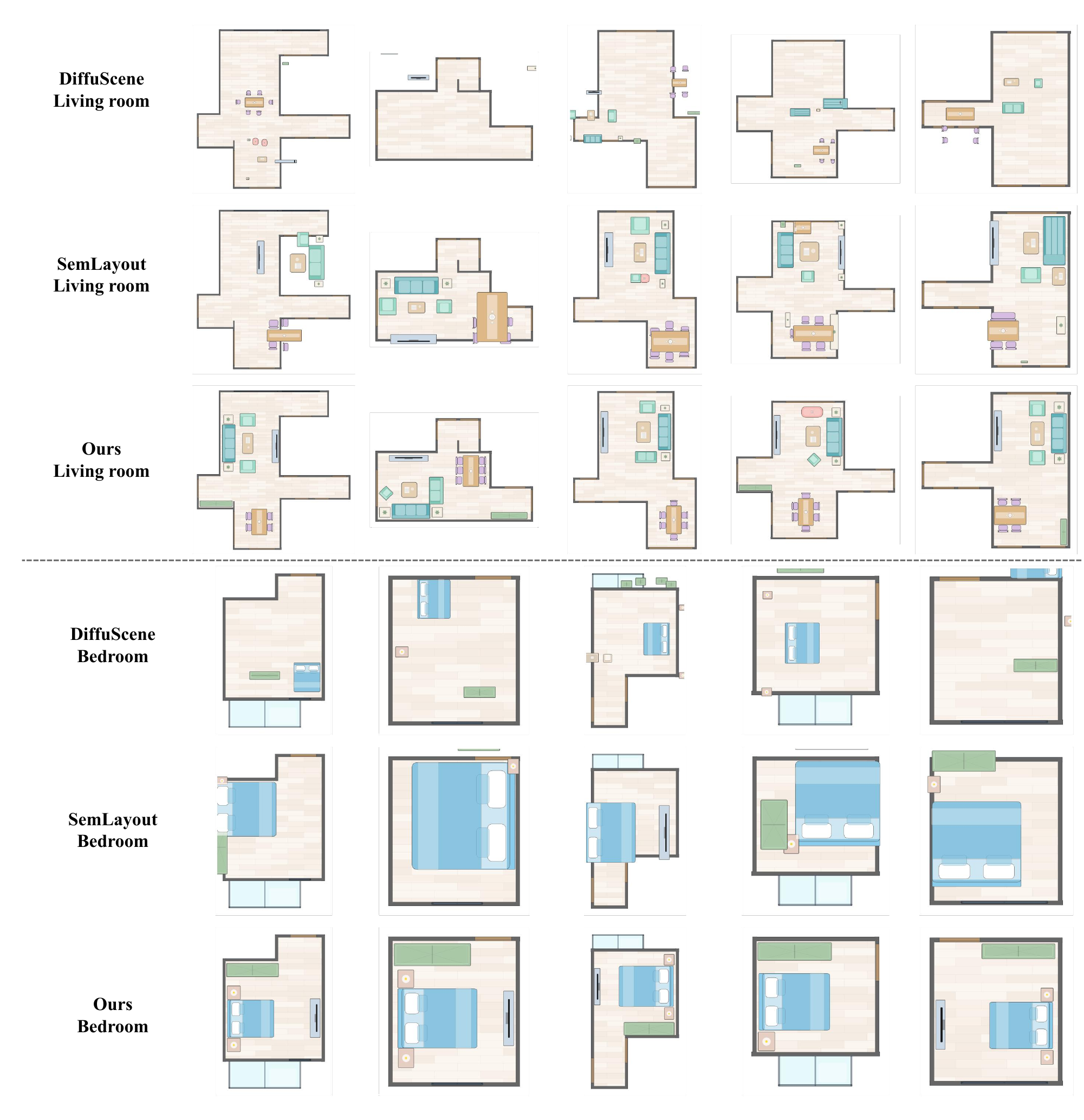}
    \caption{Qualitative comparison of DiffuScene, SemLayout, and our method on representative living room and bedroom scenes. The examples include rooms with highly irregular boundaries and high furniture density. Baseline methods exhibit typical failure modes such as out-of-bound object placement (objects outside the room are not rendered), passage blocking, object overlap, and improper object orientation, whereas our method produces compact and structurally valid layouts that better respect architectural constraints.}
    \label{fig:qualitative_comparison}
\end{figure*}

Figure~\ref{fig:qualitative_comparison} presents representative qualitative comparisons between DiffuScene, SemLayout, and our method across living room and bedroom scenes with varying structural complexity and furniture density. The selected examples highlight typical failure modes of baseline methods under complex geometric constraints.

In rooms with highly irregular boundaries, DiffuScene frequently fails to place furniture within valid room regions. Consequently, many predicted objects fall outside the room boundary and are not rendered, resulting in layouts that appear sparse and under-utilized. This behavior reflects limited capability in handling explicit structural constraints, rather than an intentional trade-off between layout density and validity. SemLayout, while capable of generating denser layouts, often violates structural constraints by blocking circulation paths, introducing object overlaps, or misaligning objects with respect to surrounding architectural elements. In contrast, our method consistently produces compact yet structurally valid layouts, preserving accessibility and respecting room boundaries even under highly constrained geometries.

These differences become more pronounced in high-density scenarios. SemLayout tends to increase furniture count at the expense of structural correctness, whereas DiffuScene struggles to maintain valid object placement as density increases. These failure modes largely stem from the lack of explicit modeling of fine-grained structural constraints during generation, which limits the ability of baseline methods to handle irregular geometries and high object density. Benefiting from an iterative evaluation-and-refinement process that explicitly enforces structural constraints, our approach achieves a better balance between space utilization and structural consistency, effectively mitigating boundary violations, collisions, and passage blocking. Overall, the qualitative results corroborate the quantitative improvements reported in Section~\ref{sec:quantitative_analysis}, demonstrating the robustness and consistency of our method under challenging structural constraints.

\subsection{User Study}
\label{sec:user_study}

To evaluate the perceptual quality and functional usability of the generated layouts beyond geometric metrics, we conducted a subjective user study involving 31 participants. We randomly sampled 30 room scenarios from the test set and generated layouts using three methods: DiffuScene~\cite{{tang2024diffuscene}}, SemLayout~\cite{sun2025semlayoutdiff}, and our Agentic Designer. Participants were presented with anonymized 2D renderings and asked to rate them on a 5-point Likert scale (1: Poor to 5: Excellent) across three dimensions: Circulation Rationality (CR), Location Rationality (LR), and Proportional Rationality (PR), as defined in Section~\ref{sec:metrics}. 

As summarized in Table~\ref{tab:user_study_sota}, our method consistently outperforms state-of-the-art baselines. Specifically, Agentic Designer achieves a CR score of 4.78, significantly surpassing DiffuScene (1.50) and SemLayout (2.83). While baselines often produce fragmented arrangements that obstruct doorways or clutter functional zones, our approach effectively preserves clear movement flows. Furthermore, our method achieves the highest ratings in Location and Proportional Rationality, confirming that the constraint-aware generation leads to layouts that are not only geometrically valid but also visually harmonious and functionally practical for real-world living.

\begin{table}[h]
    \centering
    \caption{\textbf{User Study Results.} Mean perceptual scores on a 1-5 scale (higher is better). Our method demonstrates superior performance in circulation, placement logic, and visual balance compared to baselines.}
    \label{tab:user_study_sota}
    \setlength{\tabcolsep}{12pt}
    \renewcommand{\arraystretch}{1.15}
    \begin{tabular}{lccc}
        \toprule
        \textbf{Method} & \textbf{CR} & \textbf{LR} & \textbf{PR} \\
        \midrule
        DiffuScene~\cite{tang2024diffuscene} & 1.50 & 1.32 & 1.32 \\
        SemLayout~\cite{sun2025semlayoutdiff} & 2.83 & 2.58 & 3.02 \\
        \textbf{Agentic Designer (Ours)} & \textbf{4.78} & \textbf{4.74} & \textbf{4.80} \\
        \bottomrule
    \end{tabular}
\end{table}

\subsection{Ablation Study}
\label{sec:ablation_study}

\subsubsection{Multi-Agent vs. Single-Agent Design.}

We evaluate the necessity of multi-agent collaboration by ablating the Evaluator and Refiner agents, resulting in a \textbf{Generator-Only} variant that produces layouts in a single autoregressive pass without iterative verification. This baseline is compared with the full agentic framework using both objective structural metrics and human perceptual evaluation.

Quantitative results in Table~\ref{tab:ablation_collaboration} show that removing agent collaboration leads to a clear degradation in geometric validity, with the effect being most pronounced in the Bedroom scenario under tighter spatial constraints. Specifically, the Generator-only variant exhibits a Scene Violation Rate (SVR) of 21.14\% and a Collision Score (CS) of 5.31, whereas the full framework reduces these values to 7.14\% and 3.10, respectively. This suggests that while a single Generator can produce semantically plausible layouts, it lacks the ability to reliably resolve hard geometric conflicts, which are effectively mitigated through iterative evaluation and refinement.

\begin{table}[htbp]
\centering
\caption{\textbf{Ablation Study: Effectiveness of Multi-Agent Collaboration.} Comparison between the single-agent baseline (\textbf{Generator Only}) and the full multi-agent framework. Collaboration significantly reduces violation rates (SVR) and collision scores (CS) while preserving scene complexity (ANF, OR).}
\label{tab:ablation_collaboration}
\setlength{\tabcolsep}{5pt}
\renewcommand{\arraystretch}{1.2}
\resizebox{\linewidth}{!}{
\begin{tabular}{llccccc}
\toprule
\textbf{Room Type} & \textbf{Model Variant} &
\textbf{ANF} $\uparrow$ &
\textbf{OR (\%)} $\uparrow$ &
\textbf{SVR (\%)} $\downarrow$ &
\textbf{BVC} $\downarrow$ &
\textbf{CS} $\downarrow$ \\
\midrule
\multirow{2}{*}{Living Room}
& Generator Only
& 13.15 & 21.43 & 20.00 & 0.38 & 3.42 \\
& \textbf{Agentic Designer (Full)}
& \textbf{13.16} & \textbf{21.55} & \textbf{16.85} & \textbf{0.30} & \textbf{3.27} \\
\midrule
\multirow{2}{*}{Bedroom}
& Generator Only
& 4.78 & 35.71 & 21.14 & 0.29 & 5.31 \\
& \textbf{Agentic Designer (Full)}
& \textbf{4.78} & \textbf{35.73} & \textbf{7.14} & \textbf{0.14} & \textbf{3.10} \\
\bottomrule
\end{tabular}
}
\end{table}

\begin{table}[htbp]
\centering
\caption{\textbf{User Study: Perceptual Ablation on Agent Collaboration.} Reported values are mean scores on a 1--5 scale.}
\label{tab:user_study_ablation}
\setlength{\tabcolsep}{10pt}
\renewcommand{\arraystretch}{1.12}
\begin{tabular}{lccc}
\toprule
\textbf{Configuration} & \textbf{CR} & \textbf{LR} & \textbf{PR} \\
\midrule
Generator Only & 3.63 & 3.15 & 3.70 \\
\textbf{Agentic Designer (Full)} & \textbf{4.44} & \textbf{4.39} & \textbf{4.49} \\
\bottomrule
\end{tabular}
\end{table}

Importantly, these validity improvements are achieved without simplifying the generated scenes. The Average Number of Furniture (ANF) and Occupancy Ratio (OR) remain statistically stable across configurations (e.g., ANF $\approx$ 13.15 for living rooms), indicating that the collaborative agents enhance layout validity primarily through local geometric adjustments (e.g., repositioning, reorientation, and minor rescaling) while preserving the original scene density and design intent. Consistent with these findings, a user study conducted under the same ablation setting (Table~\ref{tab:user_study_ablation}) reveals consistent perceptual declines across all criteria when agent collaboration is removed, with the largest drops observed in Location Rationality and Circulation Rationality. The full framework achieves improvements of approximately +0.7 to +0.8 across perceptual dimensions, confirming that the geometric corrections introduced by multi-agent collaboration translate directly into improved human-perceived layout quality.

\subsubsection{Impact of Progressive Consensus Mechanism}

To validate the necessity of the step-by-step verification protocol, we compare our framework against a Global Refinement variant. In this setup, the Progressive Consensus Mechanism is removed; instead, the system generates the entire furniture sequence in a single pass, followed by a one-time global evaluation and refinement of all objects. As shown in Table~\ref{tab:ablation_progressive}, replacing the progressive strategy with global refinement leads to a decline in overall geometric validity.
\begin{table}[htbp]
\centering
\caption{\textbf{Ablation Study: Progressive vs. Global Refinement.} 
Comparison between the proposed step-by-step mechanism and a global, post-hoc refinement strategy. The progressive approach achieves better overall structural validity, especially in the Bedroom scenario.}
\label{tab:ablation_progressive}
\setlength{\tabcolsep}{5pt}
\renewcommand{\arraystretch}{1.2}
\resizebox{\linewidth}{!}{
\begin{tabular}{llccccc}
\toprule
\textbf{Room Type} & \textbf{Model Variant} &
\textbf{ANF} $\uparrow$ &
\textbf{OR (\%)} $\uparrow$ &
\textbf{SVR (\%)} $\downarrow$ &
\textbf{BVC} $\downarrow$ &
\textbf{CS} $\downarrow$ \\
\midrule
\multirow{2}{*}{Living Room}
& Global Refinement
& 13.13 & 21.23 & 20.50 & 0.37 & 2.54 \\
& \textbf{Agentic Designer (Full)}
& \textbf{13.16} & \textbf{21.55} & \textbf{16.85} & \textbf{0.30} & \textbf{3.27} \\
\midrule
\multirow{2}{*}{Bedroom}
& Global Refinement
& 4.75 & 35.36 & 17.74 & 0.26 & 3.99 \\
& \textbf{Agentic Designer (Full)}
& \textbf{4.78} & \textbf{35.73} & \textbf{7.14} & \textbf{0.14} & \textbf{3.10} \\
\bottomrule
\end{tabular}
}
\end{table}

The performance degradation is most evident in the Bedroom scenario, where more compact room dimensions leave less margin for geometric adjustment, causing the Scene Violation Rate (SVR) to rise sharply from 7.14\% to 17.74\%, accompanied by a noticeable increase in collision scores (CS from 3.10 to 3.99). In the Living Room scenario, the Global variant presents a mixed outcome: while it achieves slightly lower collision scores, it fails to maintain comparable boundary adherence (20.50\% SVR vs. 16.85\%). Furthermore, other metrics also exhibit varying degrees of deterioration across both scenarios. This suggests that without step-wise error correction, the Global variant struggles to find a balanced solution that satisfies all constraints simultaneously, resulting in a suboptimal trade-off between collision avoidance and boundary compliance compared to the full Agentic Designer.

\revise{
\subsubsection{Impact of Historical Context Quality.}
To empirically validate the assumption in Eq.~(3), we isolate the impact of historical context quality on the Generator's predictions. We compare two inference settings using the same trained Generator. The two settings differ only in how the layout history is updated for subsequent steps. In the \textit{Unverified-History} setting, the Generator's raw proposal is directly appended to the context. In the \textit{Verified-History} setting, the proposal is first assessed by the Evaluator and rectified by the Refiner if necessary; this verified object is then incorporated into the context. Crucially, in both settings, quantitative metrics are computed strictly on the Generator's initial, unrefined proposals to evaluate its raw prediction capability, rather than the Refiner's immediate correction ability. By evaluating only the raw proposals in both cases, we ensure that any performance difference stems entirely from the geometric validity of the historical context guiding the Generator.

\begin{table}[htbp]
\centering
\caption{\textbf{Ablation Study: Impact of Historical Context Quality.} Comparison of the Generator's initial proposals conditioned on unverified versus verified historical contexts. Providing a verified history consistently reduces geometric violations and collisions  without sacrificing scene complexity.}
\label{tab:history_ablation}
\setlength{\tabcolsep}{5pt}
\renewcommand{\arraystretch}{1.2}
\resizebox{\linewidth}{!}{
\begin{tabular}{llccccc}
\toprule
\textbf{Room Type} & \textbf{Context Setting} & \textbf{ANF} $\uparrow$ & \textbf{OR (\%)} $\uparrow$ & \textbf{SVR (\%)} $\downarrow$ & \textbf{BVC} $\downarrow$ & \textbf{CS} $\downarrow$ \\
\midrule
\multirow{2}{*}{Living Room} 
& Unverified-History & \textbf{13.15} & \textbf{21.43} & 20.00 & 0.38 & 3.42 \\
& Verified-History & 13.01 & 21.34 & \textbf{17.97} & \textbf{0.33} & \textbf{2.19} \\
\midrule
\multirow{2}{*}{Bedroom} 
& Unverified-History & \textbf{4.78} & 35.71 & 21.14 & 0.29 & 5.31 \\
& Verified-History & \textbf{4.78} & \textbf{35.94} & \textbf{17.38} & \textbf{0.25} & \textbf{3.14} \\
\bottomrule
\end{tabular}
}
\end{table}

As shown in Table~\ref{tab:history_ablation}, maintaining this verified context directly translates into higher-quality predictions, consistent with the hypothesis in Eq.~(3). When guided by a \textit{Verified-History}, the Generator's raw proposals exhibit fewer geometric errors while maintaining comparable scene complexity (ANF and OR). Specifically, SVR/BVC/CS decrease from 20.00\%/0.38/3.42 to 17.97\%/0.33/2.19 in living rooms, and from 21.14\%/0.29/5.31 to 17.38\%/0.25/3.14 in bedrooms. This confirms that the Progressive Consensus Mechanism improves layout generation not merely by retrospectively fixing errors, but by proactively providing a clean, valid history that prevents error accumulation in subsequent steps.
}

\revise{
\subsection{Convergence and Iteration Analysis of the Refiner}
\label{sec:convergence_analysis}

The Progressive Consensus Mechanism relies on an iterative evaluation-refinement loop to enforce geometric validity. To manage computational overhead and ensure algorithmic stability, this loop is bounded by a hyperparameter: the maximum number of refinement iterations per object ($N_{max}$). Specifically, the loop terminates either when the Evaluator detects no geometric violations or when $N_{max}$ is reached, at which point the most recent state is committed. In our primary evaluations, we set the default to $N_{max}=1$ to prioritize inference efficiency while preserving high layout quality. To further investigate the convergence dynamics of the Refiner under relaxed iteration limits, we conduct an extended analysis by varying $N_{max} \in \{1, 2, 3, 4\}$ on the InStruct test set. The results are summarized in Table~\ref{tab:iteration_analysis}.
}

\begin{table}[htbp]
\centering
\caption{\textbf{\revise{Convergence Analysis with Varying Maximum Iterations ($N_{max}$).}} \revise{The average refinement steps per object plateaus rapidly, indicating fast convergence without oscillatory corrections.}}
\label{tab:iteration_analysis}
\setlength{\tabcolsep}{4pt}
\renewcommand{\arraystretch}{1.2}
\resizebox{\linewidth}{!}{
\begin{tabular}{llcccccc}
\toprule
\textbf{Room} & \textbf{$N_{max}$} & \textbf{Avg. Steps} & \textbf{ANF} $\uparrow$ & \textbf{OR (\%)} $\uparrow$ & \textbf{SVR (\%)} $\downarrow$ & \textbf{BVC} $\downarrow$ & \textbf{CS} $\downarrow$ \\
\midrule
\multirow{4}{*}{Living}
& 1 & 0.04 & 13.16 & 21.55 & 16.85 & 0.30 & 3.27 \\
& 2 & 0.05 & 13.01 & 21.33 & 10.04 & 0.24 & 1.63 \\
& 3 & 0.07 & 13.01 & 21.34 & 8.63 & 0.22 & 1.73 \\
& 4 & 0.08 & 13.02 & 21.34 & 8.43 & 0.22 & 1.61 \\
\midrule
\multirow{4}{*}{Bedroom}
& 1 & 0.07 & 4.78 & 35.73 & 7.14 & 0.14 & 3.10 \\
& 2 & 0.09 & 4.77 & 35.85 & 5.17 & 0.11 & 2.94 \\
& 3 & 0.10 & 4.77 & 35.83 & 5.31 & 0.12 & 2.72 \\
& 4 & 0.10 & 4.77 & 35.84 & 4.74 & 0.11 & 2.84 \\
\bottomrule
\end{tabular}
}
\end{table}

\revise{
As shown in Table~\ref{tab:iteration_analysis}, expanding the refinement allowance from $N_{max}=1$ to $N_{max}=2$ yields consistent improvements in structural validity. For instance, the Scene Violation Rate (SVR) in living rooms decreases from 16.85\% to 10.04\%, and the Collision Score (CS) is nearly halved. Furthermore, as $N_{max}$ is further increased to 3 and 4, the evaluation metrics stabilize, and the average number of executed refinement steps per object naturally plateaus (e.g., converging to 0.08 for living rooms and 0.10 for bedrooms).
}

\revise{
This stabilization serves as direct evidence of the system's strong convergence properties. If the Refiner were prone to oscillatory corrections, it would merely alternate between invalid states without resolving the underlying spatial conflicts. Consequently, this behavior would cause the system to continuously fail the Evaluator's checks, forcing it to exhaust the iteration budget. Under such circumstances, the average step count would increase proportionally with $N_{max}$. In contrast, the observed early plateau demonstrates that when a geometric violation occurs, the Refiner successfully rectifies it within one or two targeted adjustments. Once corrected, the object's placement immediately passes the Evaluator's validation, effectively terminating the iterative loop without falling into redundant cycles.
}

\revise{
\subsection{Inference Latency and Efficiency Analysis}
\label{sec:latency_analysis}
We report the average inference latency on the InStruct test set to quantify the computational overhead of the progressive multi-agent loop. For comparison, we evaluate a one-shot Generator baseline. Unlike the sequential iterative ablation discussed previously, this baseline predicts the complete furniture set in a single pass without intermediate evaluation or refinement. Thus, it serves as a suitable baseline to strictly isolate the latency cost introduced by the iterative verification-and-refinement process.

\begin{table}[htbp]
\centering
\caption{\textbf{Inference Latency Comparison.} Average end-to-end inference time per scene on the InStruct test set. The One-shot Generator directly predicts the complete furniture set in a single pass, while Agentic Designer performs progressive multi-agent verification and refinement.}
\label{tab:latency}
\setlength{\tabcolsep}{7pt}
\renewcommand{\arraystretch}{1.2}
\resizebox{0.82\linewidth}{!}{
\begin{tabular}{llc}
\toprule
\textbf{Room Type} & \textbf{Model Variant} & \textbf{Avg. Inference Time (s)} \\
\midrule
\multirow{2}{*}{Living Room}
& One-shot Generator & 9.87 \\
& Agentic Designer & 23.97 \\
\midrule
\multirow{2}{*}{Bedroom}
& One-shot Generator & 3.86 \\
& Agentic Designer & 8.62 \\
\bottomrule
\end{tabular}
}
\end{table}

For a fair and realistic latency comparison, all scenes are processed sequentially with a batch size of 1 in the same NVIDIA RTX 4090 GPU environment. As shown in Table~\ref{tab:latency}, Agentic Designer requires 23.97 seconds per living-room scene and 8.62 seconds per bedroom scene, compared with 9.87 and 3.86 seconds for the one-shot Generator baseline, respectively. The results indicate that the multi-agent loop increases inference time by approximately $2.2\times$--$2.4\times$. This overhead is primarily attributed to the additional evaluation and refinement steps, whereas the token generation budget remains comparable to that of the one-shot baseline. However, this increased latency is justified by improvements in overall layout quality. While single-pass generation is fast, the iterative multi-agent loop is crucial for satisfying strict structural and spatial constraints.
}

\revise{
\section{Limitation and Future Work.}
\label{sec:limitation}
While Agentic Designer significantly outperforms existing methods in structural adherence and collision avoidance, it is not without limitations. As illustrated in Figure~\ref{fig:failureCase}, highly irregular floor plans with deep external recesses or concave boundaries may still introduce topological ambiguity. In particular, the model may occasionally misinterpret nook-like external recesses as valid interior regions for furniture placement. Although the proposed multi-agent framework reduces the frequency and severity of such failures compared with baseline methods, this failure case indicates that reasoning solely over coordinate-based textual representations may be insufficient for perceiving complex 2D topology.

\begin{figure}[htbp] 
    \centering
    \includegraphics[width=\linewidth]{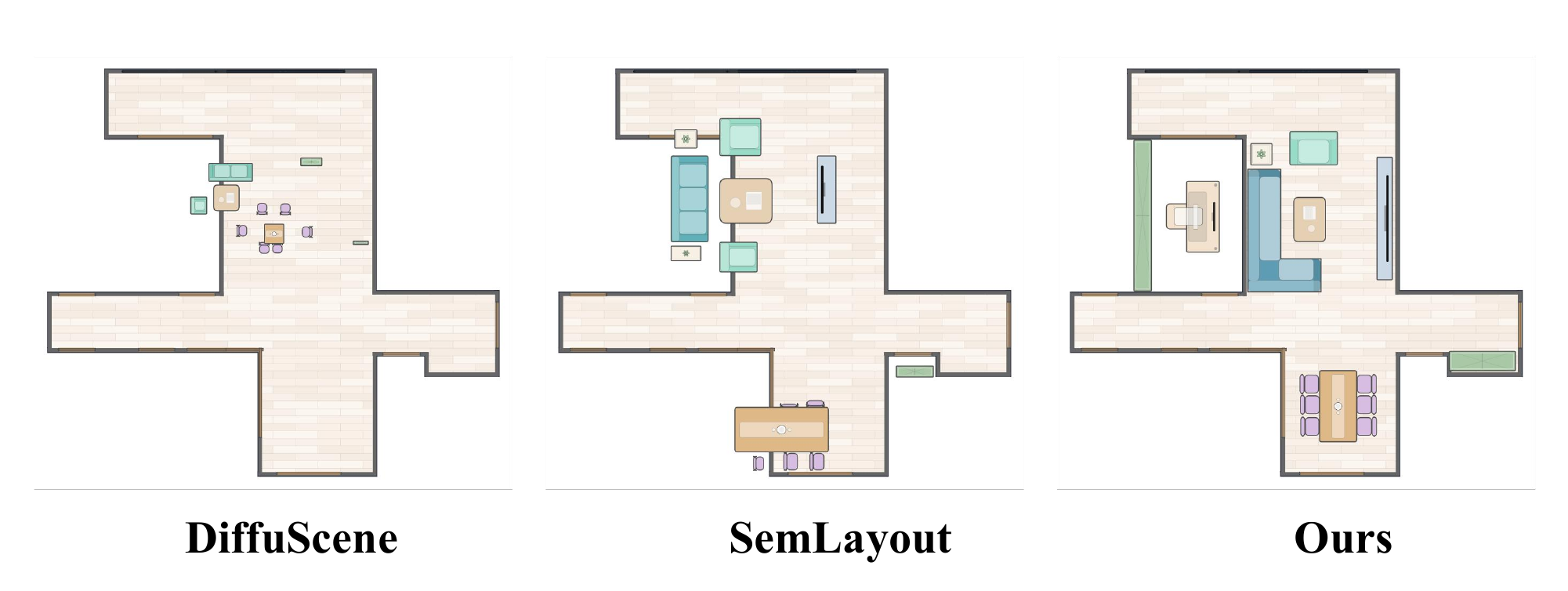}
    \caption{\textbf{Failure case analysis on highly irregular geometries.} In rooms with deep external recesses, the model may occasionally place furniture in non-interior zones due to topological ambiguity. However, compared to DiffuScene and SemLayout, Agentic Designer maintains significantly better structural coherence and minimizes such occurrences.}
    \label{fig:failureCase}
\end{figure}

This limitation is partly due to the reliance of the current Evaluator on coordinate-level geometric checks. While such checks are effective for detecting local violations, including boundary penetration, object collisions, and orientation errors, they are less reliable in determining whether a candidate placement lies in the connected interior region or in an external recess that merely appears locally valid. To address this issue, future work will incorporate visual grounding into the agentic framework. A Vision-Language Model (VLM) or a dedicated 2D spatial encoder will be introduced to extract topology-aware features from room masks, boundary contours, or occupancy maps. Serving as an auxiliary verification signal, these features will enable the Evaluator to reason more explicitly about interior-exterior relationships and global connectivity before a furniture placement is accepted. By integrating this topology-aware verification into the progressive evaluation-and-refinement loop, the framework is expected to better distinguish valid interior alcoves from invalid external recesses, thereby reducing boundary-related failures in layouts with complex and highly irregular room boundaries.
}

\section{Conclusion}
\label{sec:conclusion}
In this paper, we introduced Agentic Designer, a progressive multi-agent framework tailored for structure-aware interior layout generation. Driven by a Progressive Consensus Mechanism, our approach decomposes the complex design task into collaborative stages of proposal, evaluation, and refinement, effectively overcoming the limitations of one-shot generation methods and ensuring rigorous adherence to architectural constraints.
To standardize evaluation in this field, we also established InStruct, a comprehensive benchmark that integrates a large-scale, explicitly annotated dataset with an adapted version of 3D-FRONT and a suite of specialized structural metrics.
Our experiments on this benchmark demonstrate that the collaborative synergy between the Generator, Evaluator, and Refiner significantly reduces structural violations and collision rates, producing layouts that are both functionally plausible and geometrically coherent. We envision that Agentic Designer and the InStruct benchmark will serve as foundational resources for future research, fostering the development of more robust, controllable, and standardizable automated design tools.

\bibliographystyle{IEEEtran}
\bibliography{references}
\vspace{-20pt}
\begin{IEEEbiography}[{\includegraphics[width=1in,height=1.25in,clip,keepaspectratio]{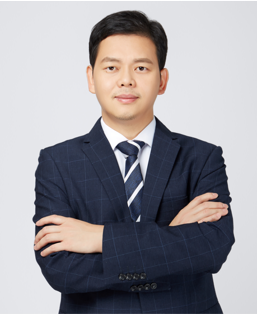}}]{Zhijing Yang} received the B.S and Ph.D. degrees from the Mathematics and Computing Science, Sun Yat-sen University, Guangzhou China, in 2003 and 2008, respectively. He was a Visiting Research Scholar in the School of Computing, Informatics and Media, University of Bradford, U.K, between July-Dec, 2009, and a Research Fellow in the School of Engineering, University of Lincoln, U.K, between Jan. 2011 to Jan. 2013. He is currently a Professor and Vice Dean at the School of Information Engineering, Guangdong University of Technology, China. He has published over 80 peer-reviewed journal and conference papers, including IEEE T-CSVT, T-MM, T-GRS, PR, etc. His research interests include machine learning and pattern recognition. 
\end{IEEEbiography}
\vspace{-15pt}

\begin{IEEEbiography}[{\includegraphics[width=1in,height=1.25in,clip,keepaspectratio]{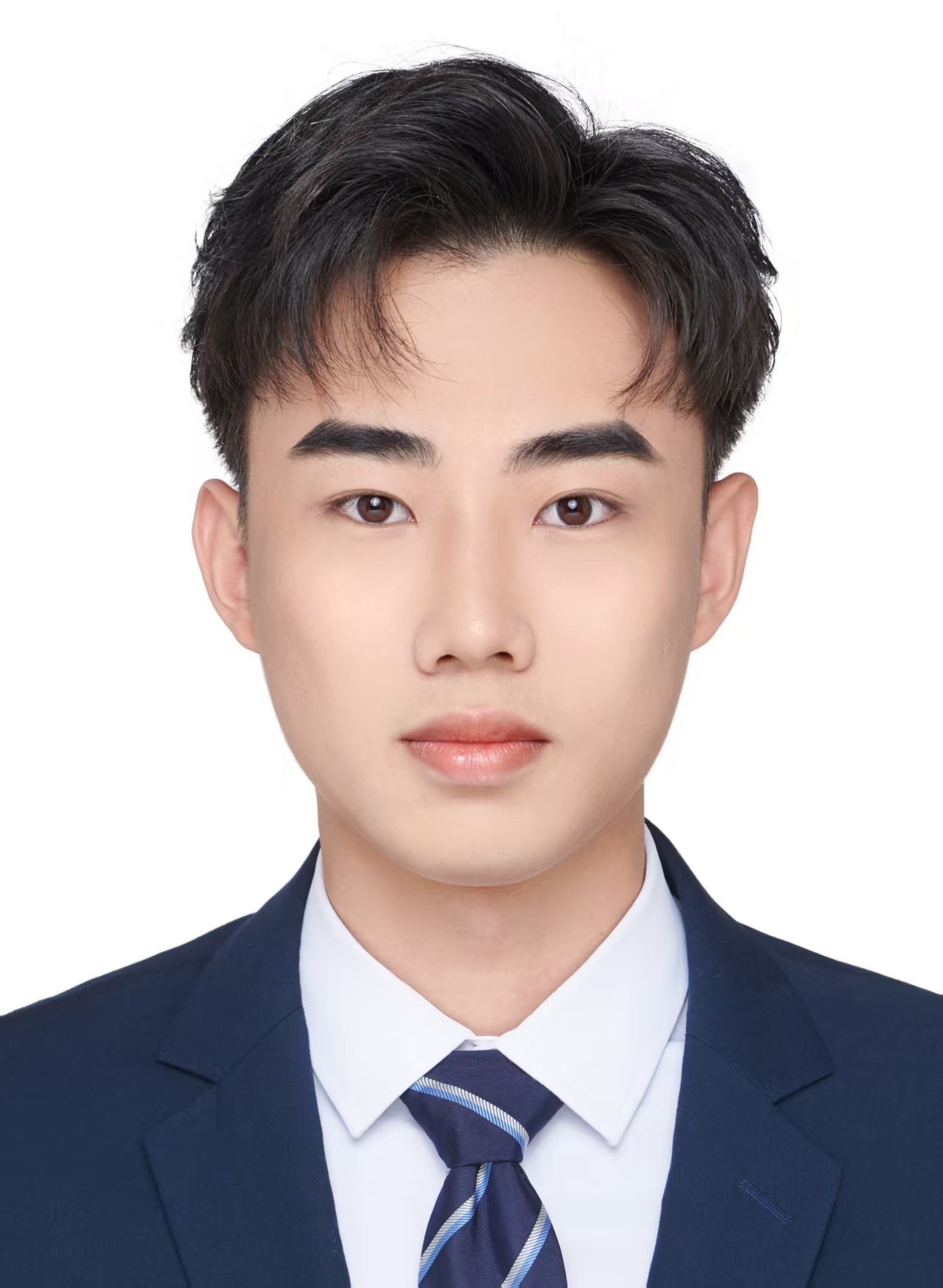}}]{Haocheng Lin} received the bachelor's degree from Guangdong University of Technology in 2024. He is currently pursuing the master's degree with the School of Information Engineering, Guangdong University of Technology, under the supervision of Prof. Zhijing Yang. His research interests include multi-agent systems and generative design. \end{IEEEbiography}
\vspace{-15pt}

\begin{IEEEbiography}[{\includegraphics[width=1in,height=1.25in,clip,keepaspectratio]{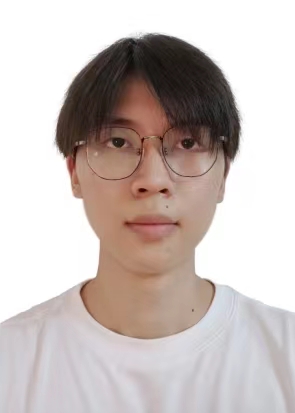}}]{Zhihua Xu} received the B.E. degree from the School of Computer Science and Technology and the M.S. degree from the School of Information Engineering, both from Guangdong University of Technology, Guangzhou, China. He is currently pursuing the Ph.D. degree with the School of Computer Science and Engineering, Sun Yat-sen University, Guangzhou, China. His research interests include Multimodal Large Language Models (MLLMs) and AI for Design. He has authored papers in top-tier conferences and journals, including ACM MM and IEEE TMM. \end{IEEEbiography}

\vspace{-15pt}
\begin{IEEEbiography}[{\includegraphics[width=1in,height=1.25in,clip,keepaspectratio]{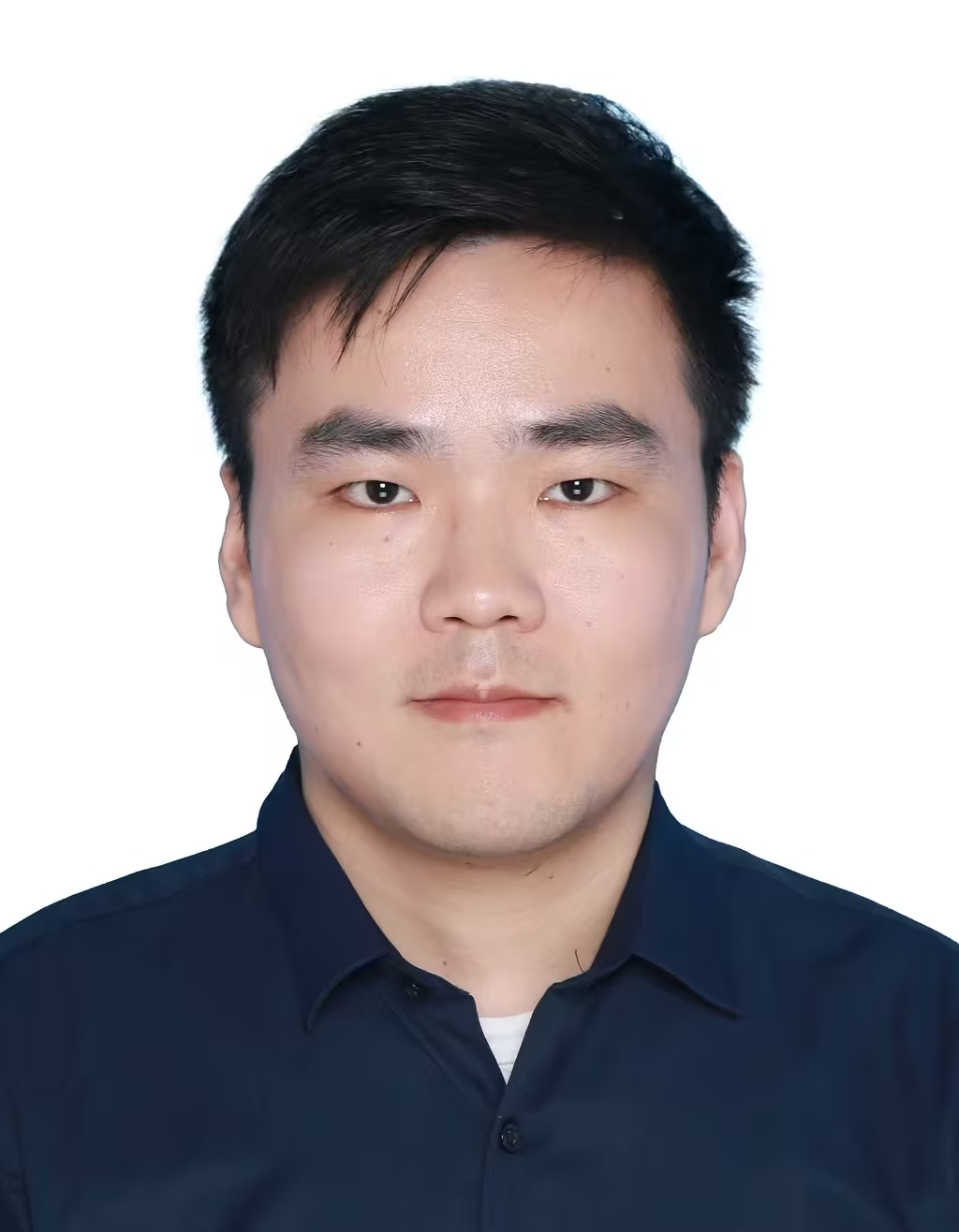}}]{Haojie Li} is currently a Ph.D. candidate at South China University of Technology. He received the M.S. degree from South China University of Technology and the B.S. degree from Shantou University. His research interests include controllable image generation and digital human generation. \end{IEEEbiography}
\vspace{-30pt}

\begin{IEEEbiography}[{\includegraphics[width=1in,height=1.25in,clip,keepaspectratio]{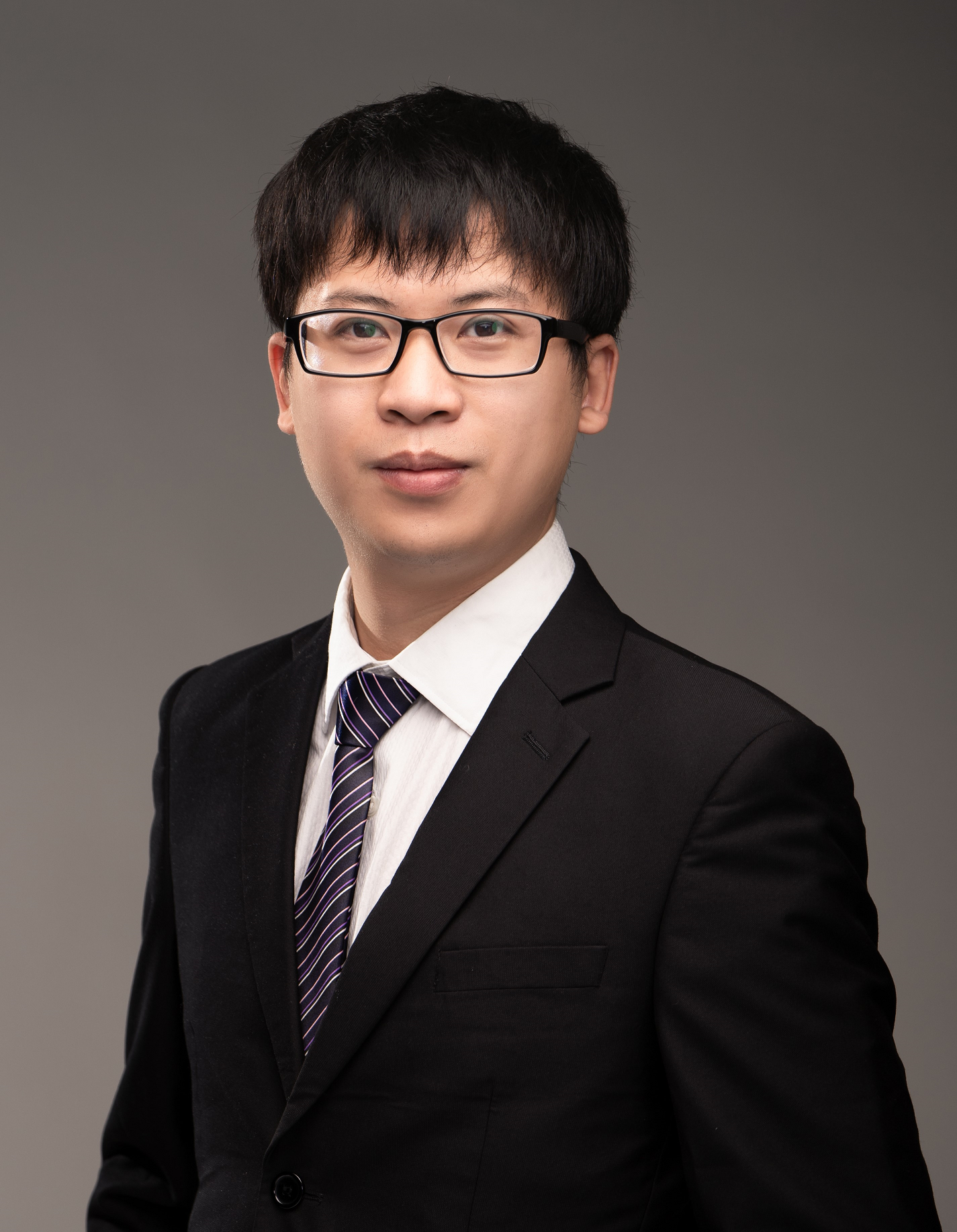}}]{Keze Wang} is nationally recognized as the Distinguished Young Scholars of the National Natural Science Foundation of China (Overseas), currently serving as an Associate Professor at the School of Computer Science, Sun Yat-sen University, and a doctoral supervisor. He holds two Ph.D. degrees, one from Sun Yat-sen University (2017) and another from the Hong Kong Polytechnic University (2019). In 2018, he worked as a postdoctoral researcher at the University of California, Los Angeles, and returned to Sun Yat-sen University in 2021 as part of the ``Hundred Talents Program''. He has focused on reducing deep learning's dependence on training samples and mining valuable information from massive unlabeled data, proposing fundamental learning paradigms, e.g., long-term self-learning and pseudo-label learning mechanisms. This has led to the gradual construction of a theoretical and methodological system for vision computing and reasoning. He has published nearly 30 papers in top-tier journals and conferences, including iScience, T-PAMI, T-NNLS, CVPR, and ICCV, with 12 papers as the first or corresponding author. His works have been cited approximately 2223 times on Google Scholar, and his has three ESI highly cited papers. He holds five patents and has received the 2018 Wu Wenjun AI Science and Technology Award, the 2019 Outstanding Doctoral Dissertation Award, and a nomination for the 2022 AI 2000 Most Influential Scholar Award. \end{IEEEbiography}
\vspace{-30pt}

\begin{IEEEbiography}[{\includegraphics[width=1in,clip]{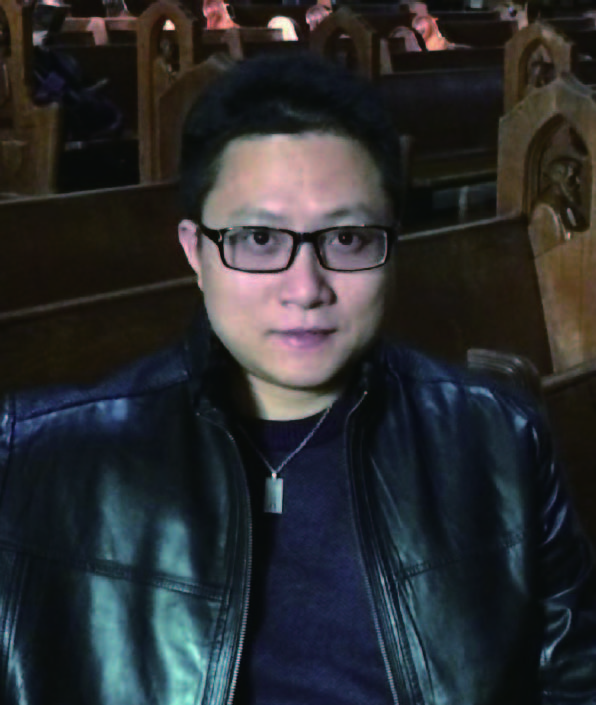}}]{Liang Lin} (Fellow, IEEE) is a full professor at Sun Yat-sen University. From 2008 to 2010, he was a postdoctoral fellow at the University of California, Los Angeles. From 2016--2018, he led the SenseTime R\&D teams to develop cutting-edge and deliverable solutions for computer vision, data analysis and mining, and intelligent robotic systems. He has authored and co-authored more than 100 papers in top-tier academic journals and conferences (e.g., 15 papers in TPAMI and IJCV and 60+ papers in CVPR, ICCV, NIPS, and IJCAI). He has served as an associate editor of IEEE Trans. Human-Machine Systems, The Visual Computer, and Neurocomputing and as an area/session chair for numerous conferences, such as CVPR, ICME, ACCV, and ICMR. He was the recipient of the Annual Best Paper Award by Pattern Recognition (Elsevier) in 2018, the Best Paper Diamond Award at IEEE ICME 2017, the Best Paper Runner-Up Award at ACM NPAR 2010, Google Faculty Award in 2012, the Best Student Paper Award at IEEE ICME 2014, and the Hong Kong Scholars Award in 2014. He is a Fellow of IEEE, IAPR, and IET. \end{IEEEbiography}
\vspace{-30pt}

\begin{IEEEbiography}[{\includegraphics[width=1in,height=1.25in,clip,keepaspectratio]{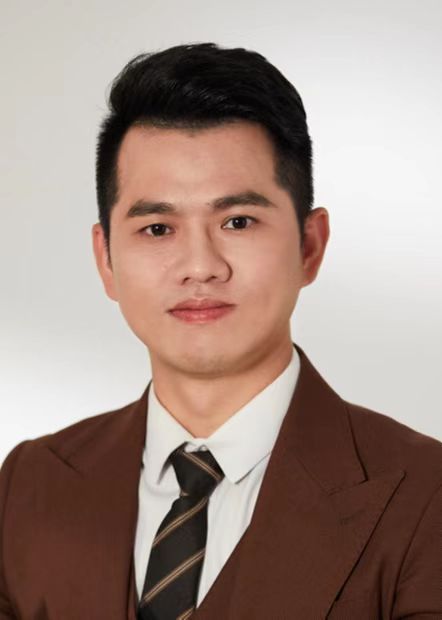}}]{Tianshui Chen} received a Ph.D. degree in computer science at the School of Data and Computer Science Sun Yat-sen University, Guangzhou, China, in 2018. Prior to earning his Ph.D, he received a B.E. degree from the School of Information and Science Technology in 2013. He is currently an associate professor at the Guangdong University of Technology. His current research interests include artificial intelligence, multimodal large models, and generative AI. He has authored and co-authored more than 60 papers published in top-tier academic journals and conferences, including T-PAMI, IJCV, T-NNLS, T-IP, T-MM, CVPR, ICCV, AAAI, IJCAI, ACM MM, etc. He has served as a reviewer for numerous academic journals and conferences. He was the recipient of the Best Paper Diamond Award at IEEE ICME 2017. \end{IEEEbiography}
\vspace{-20pt}

\end{document}